\documentclass[10pt,twocolumn,letterpaper]{article}

\usepackage{cvpr}
\usepackage{times}
\usepackage{epsfig}
\usepackage{graphicx}
\usepackage{amsmath}
\usepackage{amssymb}
\usepackage[ruled]{algorithm2e}
\usepackage{multirow}
\usepackage{indentfirst}
\usepackage{graphicx}
\usepackage{float}
\usepackage{dirtree}
\usepackage{amsmath,amssymb}
\usepackage{setspace}
\usepackage{mathtools}
\usepackage{float}
\usepackage[normalem]{ulem}
\usepackage{array}
\usepackage{epstopdf}
\usepackage{amsmath}
\usepackage{booktabs}
\usepackage{cite}
\usepackage{times}
\usepackage{epsfig}
\usepackage{graphicx}
\usepackage{amsmath}
\usepackage{amssymb}
\usepackage{caption}
\usepackage{color}
\usepackage{subfig}
\usepackage{lipsum}
\usepackage{booktabs}
\usepackage{bm}
\usepackage{multicol}
\usepackage{multirow}
\usepackage{bbding}
\usepackage{array}
\usepackage{authblk}

\newcommand{\ignorethis}[1]{}

\usepackage[pagebackref=true,breaklinks=true,letterpaper=true,colorlinks,bookmarks=false]{hyperref}

 \cvprfinalcopy 

\def\cvprPaperID{0000} 

\ifcvprfinal\fi
\begin{document}


\title{An End-to-end Method for Producing Scanning-robust Stylized QR Codes}

\author[1]{Hao Su}
\author[1]{Jianwei Niu\thanks{C.C@university.edu}}
\author[1]{Xuefeng Liu}
\author[1]{Qingfeng Li}
\author[1]{Ji Wan}
\author[2]{Mingliang Xu}
\author[1]{Tao Ren}

\affil[1]{BUAA-Lab of Distributed and Mobile Computing, Beihang University, Beijing, China}
\affil[2]{School of Computer and Artificial Intelligence, Zhengzhou University, Zhengzhou, China}
\affil[*]{The corresponding author. \tt\small niujianwei@buaa.edu.cn }

\renewcommand\Authands{ and }

\twocolumn[{%
\renewcommand\twocolumn[1][]{#1}%
\maketitle
\begin{center}
\vspace{-0.8cm}
    \centering
    \includegraphics[width=6.8 in]{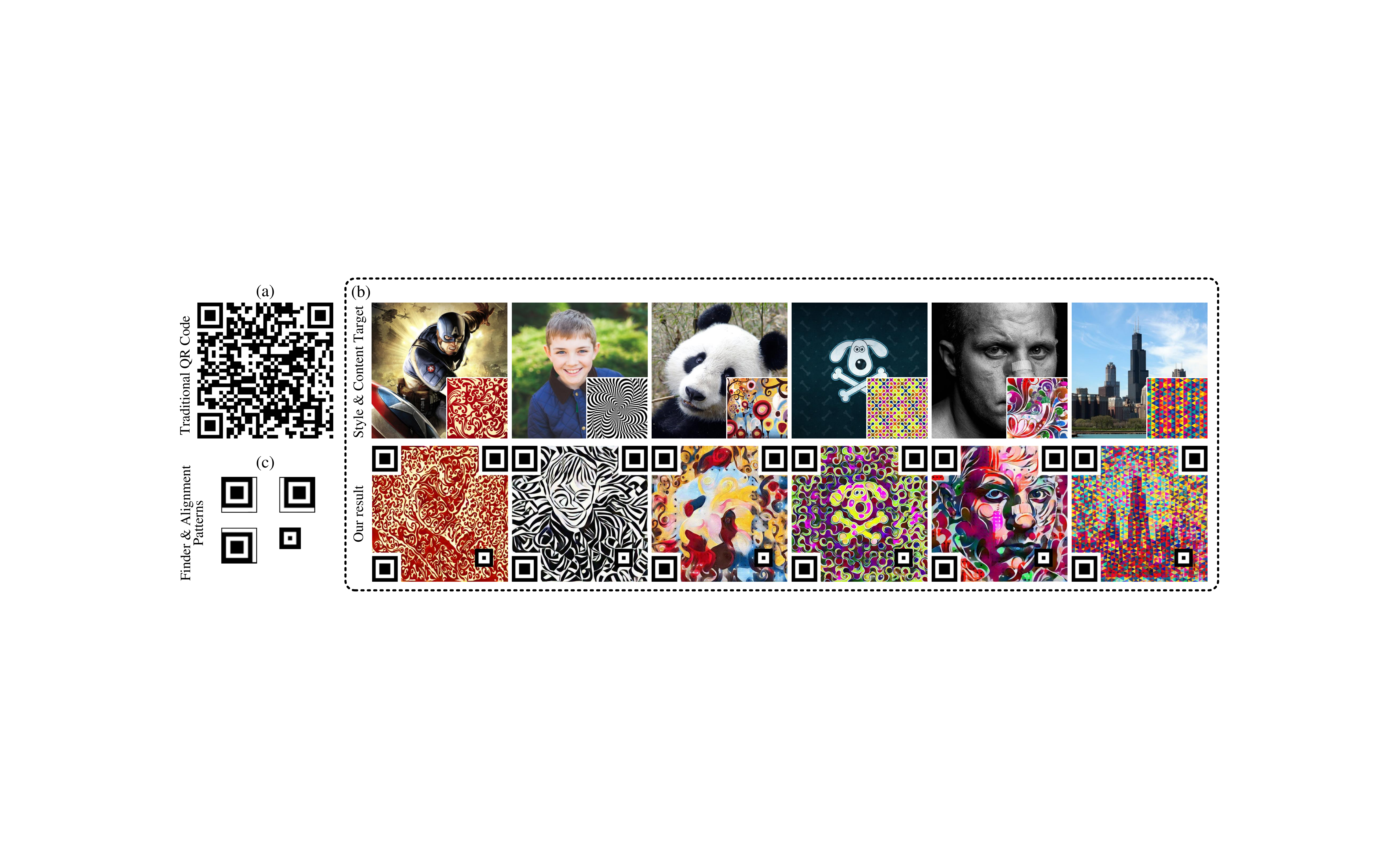}
   \vspace{-0.2cm}
    \captionof{figure}{(a) Traditional QR code. (b) Samples of our stylized QR codes. These codes combine the visual effect of stylization
and the functionality of QR codes, which are personalized, attractive, and scanning-robust. (c) Finder and alignment patterns are used to determine a QR code's location and angle, thus we preserve their traditional appearances. }
    \label{teaser}
\end{center}%
}]


\begin{abstract}
\vspace{-0.8cm}
\emph{Quick Response (QR) code} is one of the most worldwide used two-dimensional codes.~Traditional QR codes appear as random collections of black-and-white modules that lack visual semantics and aesthetic elements, which inspires the recent works to beautify the appearances of QR codes. However, these works adopt fixed generation algorithms and therefore can only generate QR codes with a pre-defined style. In this paper, combining the Neural Style Transfer technique, we propose a novel end-to-end method, named ArtCoder, to generate the stylized QR codes that are personalized, diverse, attractive, and scanning-robust.~To guarantee that the generated stylized QR codes are still scanning-robust, we propose a Sampling-Simulation layer, a module-based code loss, and a competition mechanism. The experimental results show that our stylized QR codes have high-quality in both the visual effect and the scanning-robustness, and they are able to support the real-world application.
\end{abstract}

\section{Introduction}

With the ubiquity of smartphones, the \emph{Quick Response (QR) code} \cite{ISO} has become one of the most-used types of two-dimensional codes, and has been popularly applied in many scenarios including social networks, mobile payments, and advertisements.~Traditional QR codes are matrix codes consisting of black-and-white squares \emph{modules} that are visual-unpleasant and meaningless to human vision [Fig.~\ref{teaser}(a)].~An appealing QR code will attract more people to scan the code and increase the link visits \cite{HF,EF,ARTUP}, which inspires the recent works to beautify the appearances of QR codes \cite{HF,VS,EF,TS,Masic,ARTUP} (i.e., endowing QR codes with visual semantics or aesthetic elements). However, these works adopt fixed generation algorithms and therefore can only generate QR codes with a pre-defined style, [Fig.~\ref{fig:relatedwork}(b)-(f)], which limits the personalized choices of users.

In this paper, employing the Neural Style Transfer (NST) technique, we propose an end-to-end method, named \emph{ArtCoder}, to generate the stylized QR codes. These stylized codes combine the visual effect of stylized images and the functionality of QR codes, which are personalized, diverse, and scanning-robust [as shown in Fig.~\ref{teaser}(b)]. Although the recent NST works (e.g.,\cite{Gatys,Markov,stylebank,mangagan,Lff}) have made great progress on stylizing images, however, for generating stylized QR codes, the big extra challenge is to guarantee the scanning-robustness of output codes after giving them art styles.~To address this issue, Xu \emph{et al}.\cite{StylizeQR} propose a two-staged method, i.e, a QR code is first stylized by a NST model \cite{Lff}, and then all error modules caused by stylization are repaired by a post-processing algorithm.~Although this method can produce the stylized QR codes, as shown in Fig.~\ref{fig:relatedwork}(h) upper row, the repaired modules are distracting and cannot be well fused with the entire image, due to the asynchrony of stylization and module repair.

Unlike existing works, our ArtCoder is the first end-to-end method to stylize an image and fuse it with the QR code message simultaneously.
Moreover, ArtCoder not only blends the black/white modules in an invisible and attractive manner [Fig.~\ref{teaser}(b)], but also preserves the scanning-robustness. To improve the performances in both scanning-robustness and visual quality, we propose three key improvements as follows.~First, we analysis the relationship between the convolutional layers and the sampling process of QR code readers, and propose a Sampling-Simulation (SS) layer to extract the encoding message of QR codes. Second, we propose a module-based \emph{code loss} to control the scanning-robustness of the stylized QR codes. Third, we propose a competition mechanism between the visual quality and the scanning-robustness to improve their performances.

The main contributions of our work are three-fold:
\vspace{-0.2cm}
\begin{itemize}\setlength{\itemsep}{-2pt}
	\item we propose a novel end-to-end method ArtCoder to generate the stylized QR codes that are personalized, diverse, and scanning-robust.
	\item we propose a Sampling-Simulation (SS) layer to extract the message of QR codes, and introduce the module-based {code loss} to preserve the scanning-robustness of the stylized QR codes.
    \item we propose a competition mechanism to guarantee the high-quality of the stylized QR codes in both scanning-robustness and visual effect.
\vspace{-0.2cm}
\end{itemize}

\section{Related Work}
\vspace{-0.2cm}
Below we summarize the related works that involve two main topics, neural style transfer, and aesthetic QR code.
\vspace{-0.1cm}
\subsection{Neural style transfer}
\vspace{-0.1cm}
The methods of Neural Style Transfer (NST) can be basically classified as parametric and non-parametric \cite{reshuff}.

\textbf{Parametric.}~Parametric methods iteratively update an initial image until the desired global statistics are satisfied. Gatys~\emph{et~al.} \cite{Gatys} pioneer the parametric NST method by employing the power of CNN and Gram matrices. Afterwards, the follow-up parametric researches have been presented to improve their performances on visual quality \cite{SeparatingNST_CVPR2018, Arbitrary_NST_CVPR2018,Stroke_NST_ECCV2018, text_CVPR2018}, generating speed \cite{stylebank,Lff,Arbitraryfast_CVPR2017,NSTCVPR2018,universalNST_NIPS2017}, and multimedia extension \cite{Coherent, NST_VIDEO_CVPR2017,StylizeQR, stereoscopicNST_CVPR2018}.

\textbf{Non-parametric.}~Non-parametric methods use a simple patch representation, and find the most similar patches by nearest neighbor search. Non-parametric NST method is pioneered by Li \emph{et al.} \cite{Markov}, and they reformulate the style transfer using the Markov Random Field (MRF), i.e., searching neural patches from the style image to match the structure of content image. Afterwards, various researchers follow the idea of patch-based matching to optimize the stylized results by de-VGG networks \cite{fastpatch}, semantic-level patch \cite{dia}, feature reshuffle \cite{reshuff}, etc.

\begin{figure}[t]
\centering
\includegraphics[width=3.2 in]{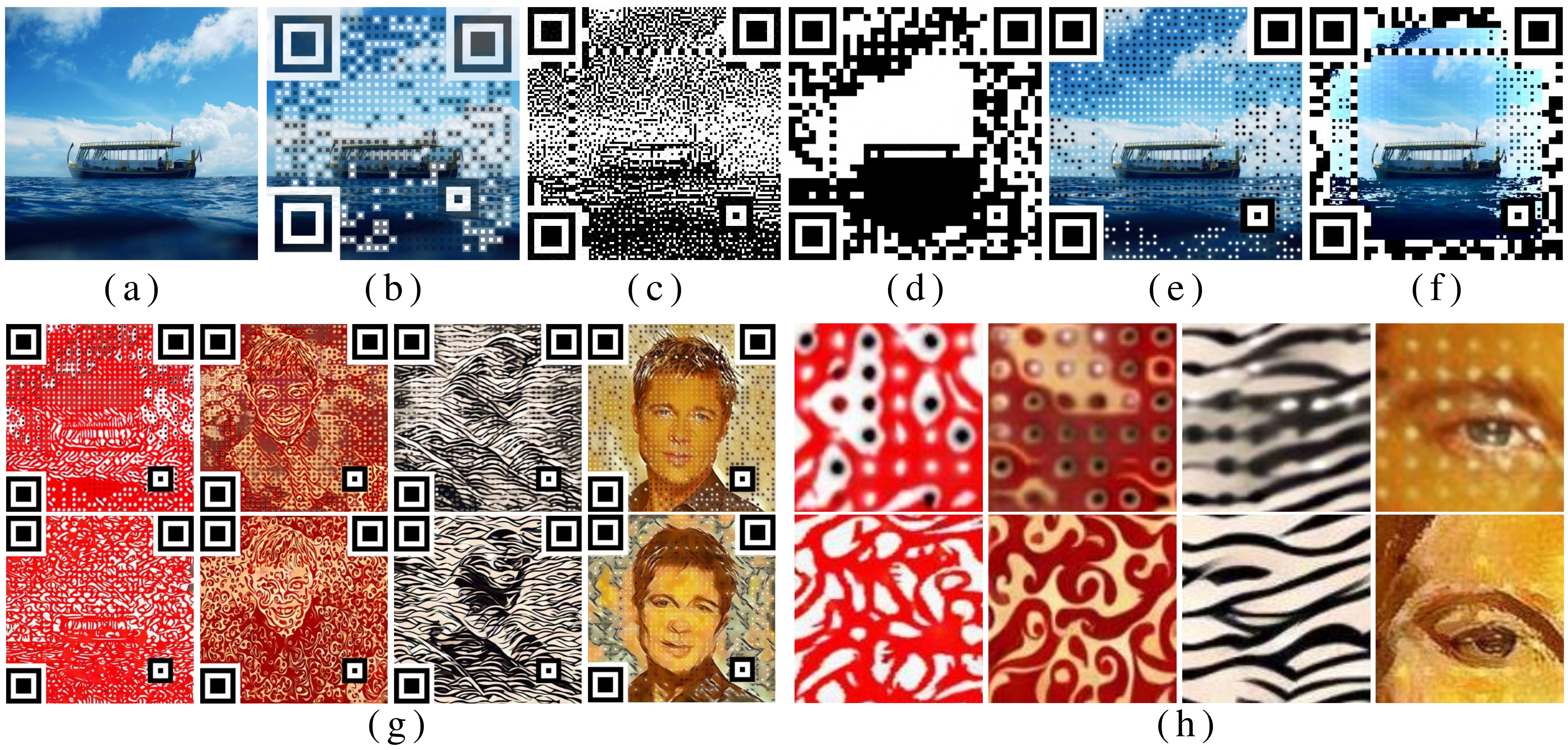}
\vspace{-0.3cm}
\caption{(a) Blended image. (b) Visualead \cite{VS}. (c) Halftone QR code \cite{HF}. (d) Qart code \cite{Cox}. (e) Artup \cite{ARTUP}. (f) Efficient QR code \cite{EF}.  (g) SEE QR code \cite{StylizeQR} (upper) and our results (bottom). (h) Enlarged view of (g).}
\vspace{-0.5cm}
\label{fig:relatedwork}
\end{figure}

\begin{figure*}[t]
\vspace{-0.4cm}
\centering
\includegraphics[width=5.5 in]{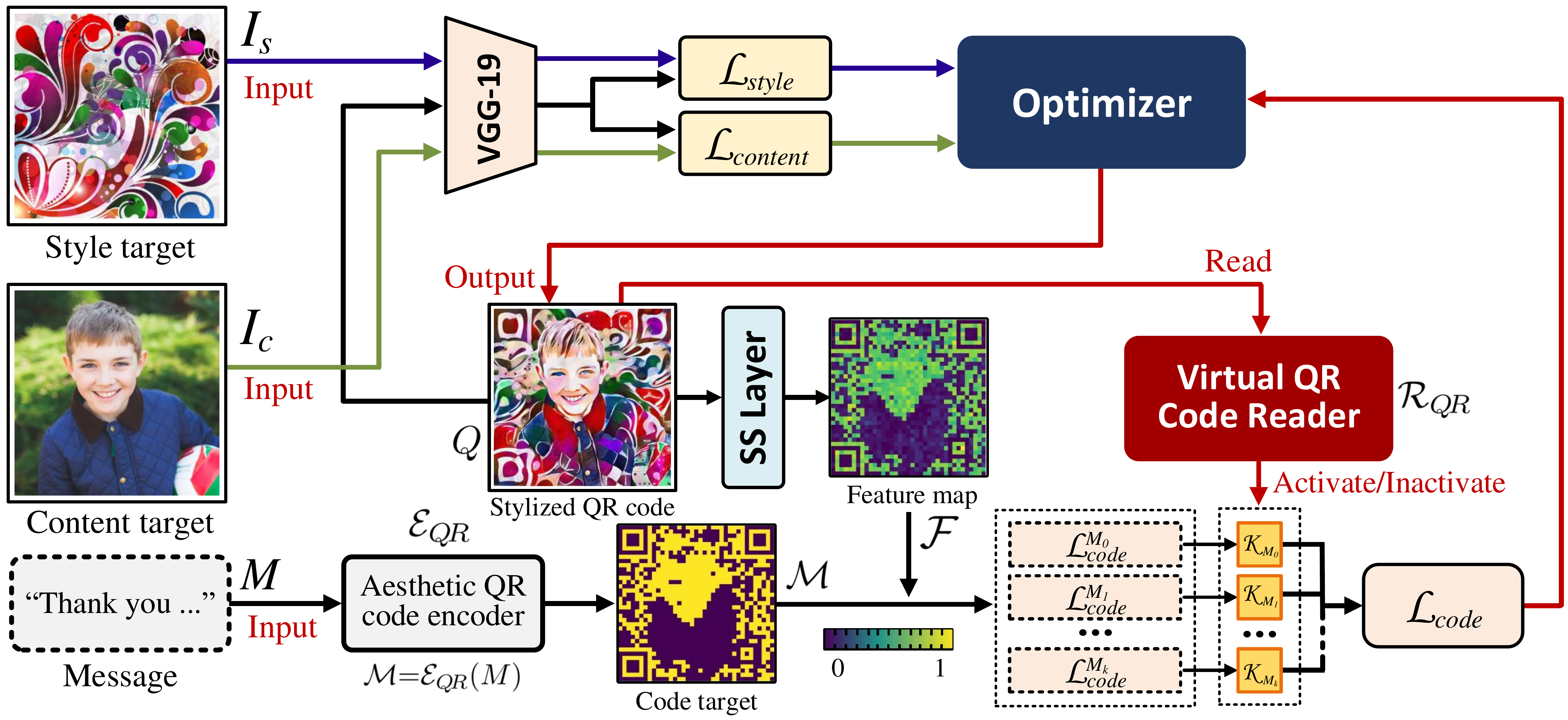}
\vspace{-0.2cm}
\caption{System Pipeline.~Given a style target image $I_s$, a content target image $I_c$, and a message $M$, our method is modeled as a function $\Psi$ to generate a stylized QR code $Q$$=$$\Psi(I_s,I_c,M)$. For visual effect, $Q$ combines the style feature of $I_s$ and the semantic content of $I_c$, and for functionality, $Q$ can be decoded to message $M$ by any standard QR code reader.
}
\vspace{-0.4cm}
\label{fig:overview}
\end{figure*}

\vspace{-0.1cm}
\subsection{Aesthetic QR code}
\vspace{-0.1cm}

Bellow we review the methods of blend-type aesthetic QR codes \cite{StylizeQR} that can blend images with QR codes (Fig.~\ref{fig:relatedwork}), and these methods are mainly based on three directions, module-deformation, module-reshuffled, and NST.

\textbf{Module-deformation.}~The idea of methods based on module-deformation is first to deform and reduce the regions of square modules, and then insert images in the saved regions, where the representative works are Visualead \cite{VS} and Halftone QR codes \cite{HF}.~Visualead \cite{VS} beautifies QR codes by deforming modules and keeping the contrast between modules and blended images [Fig.~\ref{fig:relatedwork}(b)]. Halftone QR codes \cite{HF} divide each module into 3$\times$3 submodules with keeping the color of the center sub-modules, and then make the other sub-modules to match the halftone map of the blended image [Fig.~\ref{fig:relatedwork}(c)].

\textbf{Module-reshuffle.}~Recent methods based on module-reshuffle are inspired by the pioneering work Qart code \cite{Cox} which proposes that the \emph{Gauss-Jordan Elimination Procedure} can be employed to reshuffle modules' locations to satisfy the features of blended images [Fig.~\ref{fig:relatedwork}(d)]. Afterwards, aiming at improving the visual quality of QR codes, the follow-up works design different strategies to reshuffle modules using different image features, e.g., region of interesting \cite{ARTUP}, central saliency \cite{EF}, global gray values \cite{StylizeQR}.

\textbf{NST-based method.}~Xu \emph{et al.} \cite{StylizeQR} first introduce the NST technique to generate stylized QR codes, and propose the SEE (Stylized aEsthEtic) QR codes [Fig.~\ref{fig:relatedwork}(g) upper row] that are personalized and machine-readable.~Their method address the issue that the style transfer will compromise the scanning-robustness, however, the error modules caused by stylization are repaired by a post-processing algorithm, which generates the distracting modules that cannot be well fused with the entire image [Fig.~\ref{fig:relatedwork}(h) upper row].

\section{Method}
\subsection{Overview}
 Given a style target image $I_s$, a content target image $I_c$, and a message $M$, our method is modeled as a function $\Psi$ to generate a stylized QR code $Q\!=\!\Psi(I_s,I_c,M)$. For visual effect, $Q$ combines the style feature of $I_s$ and the semantic content of $I_c$; for functionality, $Q$ can be scanned to show message $M$ by any standard QR code reader.
The total objective function $\mathcal{L}_{total}$ of $Q\!=\!\Psi(I_s,I_c,M)$ is defined as
\begin{equation}
\label{equation:totalloss}
\begin{aligned}
\mathcal{L}_{total} =& \lambda_1 \mathcal{L}_{style}(I_s,Q)+ \lambda_2 \mathcal{L}_{content}(I_c,Q) \\
 & + \lambda_3 \mathcal{L}_{code}(M,Q)
\end{aligned} ,
\end{equation}
where $\lambda_1$ to $\lambda_3$ are used to balance the multiple objectives. Style loss $\mathcal{L}_{style}$, content loss $\mathcal{L}_{content}$, and code loss $\mathcal{L}_{code}$ are minimized by the optimizer, to control the style feature, semantic content, and readability of $Q$, respectively.

As shown in Fig.~\ref{fig:overview}, the features of style and content are extracted by VGG-19 \cite{VGG16}, and the code feature is extracted by the proposed Sampling-Simulation (SS) layer. In each iteration of optimizer, a virtual QR code reader $\mathcal{R}_{Q\!R}$ will read the stylized result $Q$ to discriminate all error and correct modules.~For the $k$-th module $M_k$, if $M_k$ is error (or correct), we control an activation map $\mathcal{K}$ to activate (or inactivate) the $k$-th sub-code-loss $\mathcal{L}_{code}^{\!M_k\!}$, to optimize its robustness and may compromise the representations of style and content (or $\mathcal{L}_{style}$ and $\mathcal{L}_{content}$ will try their best to optimize the visual quality).


We will detail the losses of style and content, the SS layer, the code loss, and the virtual QR code reader in Sec.~\ref{section:styleandcontent}, Sec.~\ref{section:SSlayer}, Sec.~\ref{section:codeloss}, and Sec.~\ref{section:reader}, respectively.

\subsection{Losses of style and content}
\vspace{-0.1cm}
\label{section:styleandcontent}
The style loss $\mathcal{L}_{style}$ and the content loss $\mathcal{L}_{content}$ are not the key points of our work, thus we basically follow the literature \cite{Gatys, Lff} as
\vspace{-0.2cm}
\begin{equation}
\label{equation:stylecontentloss}
\begin{aligned}
&\mathcal{L}_{style}(I_s,Q) \! = \! \frac{1}{C_sH_sW_s} \big\| {G}[f_s(I_s)] \! - \! {G}[f_s(Q)] \big\|^2_2 \\
&\mathcal{L}_{content}(I_c,Q)\! =\! \frac{1}{C_cH_cW_c} \big\| f_c(I_c)-f_c(Q) \big\|^2_2
\end{aligned} ,
\vspace{-0.1cm}
\end{equation}

where ${G}$ indicates the \emph{Gram matrix} \cite{Gatys, Lff}, and $f_s$ (or $f_c$) is the feature map of shape $C_s$$\times$$H_s$$\times $$W_s$ (or $C_c$$\times$$H_c$$\times$$W_c$) that extracted from the $s$-th (or $c$-th) layer of the pre-trained VGG-19 network \cite{VGG16}, $s$$\in$\{relu1\_2, relu2\_2, relu3\_3, relu4\_3\}, and $c$$\in$\{relu3\_3\}.

\subsection{Sampling-Simulation layer}
\label{section:SSlayer}
\textbf{Sampling of QR codes:} for decoding QR codes, the most used project \emph{Google ZXing} \cite{ZXing} rules that a QR code reader only samples the center pixel of each module in a QR code, and then binarizes and decodes these pixels. In other words, a QR code is still readable if replacing all original square modules with smaller concentric modules.~Meanwhile, the probability of sampling correct pixels is not fixed and proportional to the module sizes, due to the external factors (e.g., camera resolution, scanning distance) when QR codes are scanned by smartphones.~To theorize this point, \cite{StylizeQR, ARTUP} propose that the pixels closer to the module center have a higher probability to be sampled, and the probability follows the {Gaussian distribution} $\mathcal{G}$ as
\vspace{-0.2cm}
\begin{equation}
\label{equation:Gauss}
\mathcal{G}_{M_k(i,j)}=\frac{1}{2\pi \sigma^2}e^{-\frac{i^2 + j^2}{2\sigma^2}},
\end{equation}
where $(i, j)$ is the coordinate of a pixel in module $M_k$, and the origin at the module center, and $\mathcal{G}_{M_k}\!$$(i,j)$ indicates the probability of sampling the pixel $(i,j)$. 

\begin{figure}[t]
\centering
\vspace{-0.4cm}
\includegraphics[width=2.7 in]{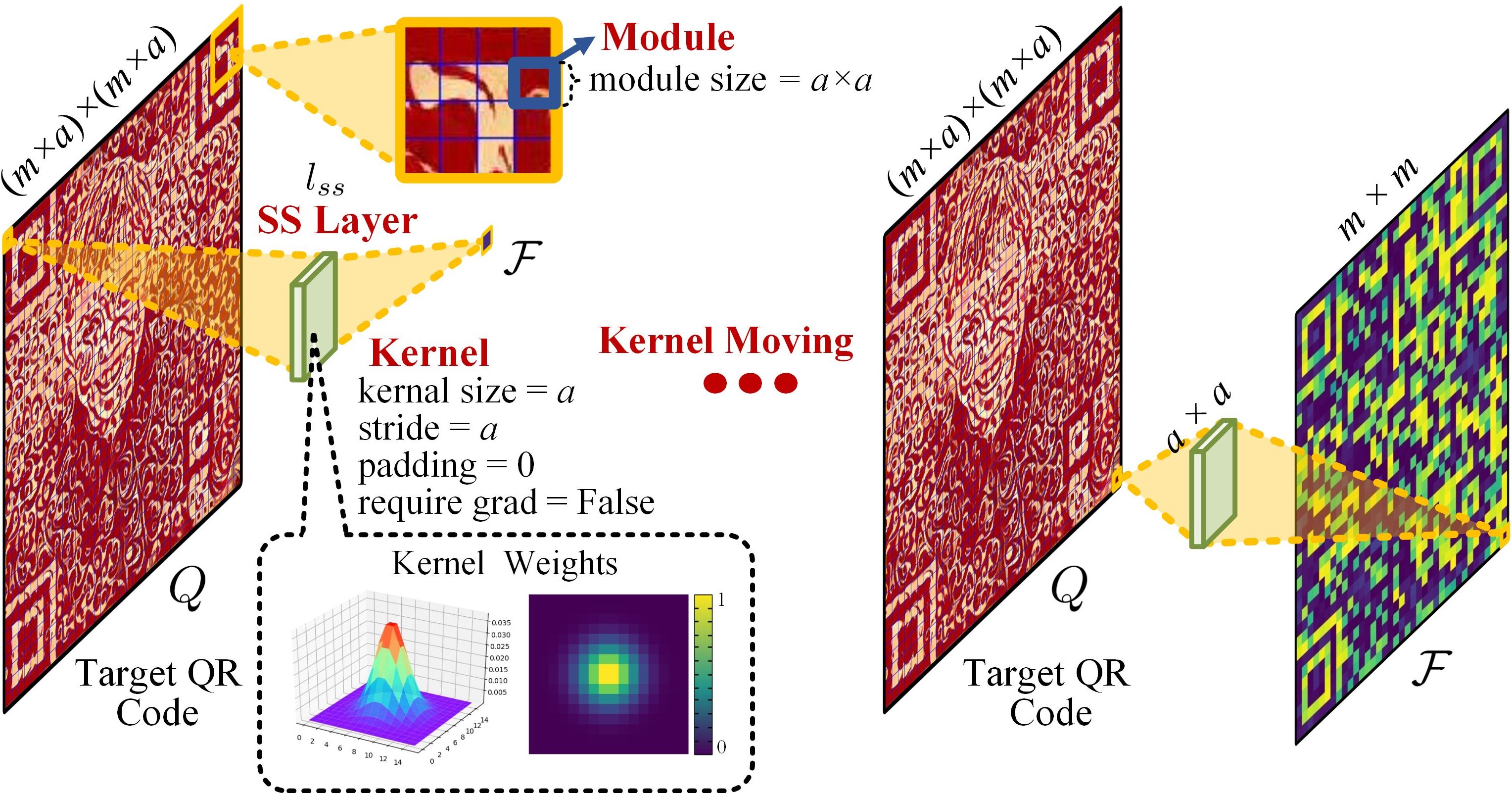}
\vspace{-0.1cm}
\caption{Framework of Sampling-Simulation (SS) layer $l_{ss}$. For target QR code $Q$ that consists of $m$$\times$$m$ modules of size $a$$\times$$a$ pixels, $l_{ss}$ extracts an $m$$\times$$m$ feature map $\mathcal{F}$ from $Q$, and $\mathcal{F}$ indicates the sampled colors of all modules in $Q$. The kernel weight of $l_{ss}$ follows the Gaussian distribution as eq.(\ref{equation:Gauss}), since the pixels closer to module center are more important for the scanning-robustness. }
\label{fig:lss}
\vspace{-0.4cm}
\end{figure}

\textbf{Sampling Simulation:} If using a conv layer to simulate the sampling process of a QR code reader, we can control the robust of QR code by the losses back propagation. To achieve this goal, we analysis the relationship between convolution and sampling, and further design the Sampling-Simulation (SS) conv layer $l_{ss}$.

As shown in Fig.~\ref{fig:lss}, for a stylized QR code $Q$ that consists of $m$$\times$$m$ modules of size $a$$\times$$a$, $l_{ss}$ is designed to have kernel size $a$, stride $a$, padding 0, and the kernel weights are fixed to follow the Gaussian weight as eq.(\ref{equation:Gauss}). When we input $Q$ to $l_{ss}$, the kernel will convolve each module of $Q$ once, and output an $m$$\times $$m$ feature map $\mathcal{F}$$=$$l_{ss}(Q)$, and $\mathcal{F}$ indicates the sampled results of $Q$. Each bit $\mathcal{F}_{M_k} $ in $ \mathcal{F}$ is correspond to the module $M_k$ in $Q$, represented as
\vspace{-0.2cm}
\begin{equation}
\label{equation:lss}
\mathcal{F}_{M_k} =  \!\!\!\!\!\!\!\!\!  \sum_{{ \ \ \ \ (i,j) \in M_k}} \!\!\!\!\!\!\!\!\! \mathcal{G}_{M_k (i,j) }\!\cdot\! Q_{M_k (i,j)} ,
\vspace{-0.2cm}
\end{equation}
where $\mathcal{G}_{M_k (i,j)}$ defined as eq.(\ref{equation:Gauss}).

%

\subsection{Code loss and competition mechanism}
\vspace{-0.1cm}
\label{section:codeloss}
\textbf{Code loss:} the code loss $\mathcal{L}_{code}$ is based on the module of QR code $Q$, that is, we set a sub-code-loss $\mathcal{L}_{code}^{M_k}$ for each module $M_k$$\in$$Q$, and sum them up to get the total code loss $\mathcal{L}_{code}$ as
\vspace{-0.2cm}
\begin{equation}
\label{equation:codelosstotal}
\mathcal{L}_{code} \  = \!\! \sum_{M_k\in Q}  \mathcal{L}_{code}^{M_k} \ \ .
\vspace{-0.2cm}
\end{equation}

For the input message $M$, we encode $M$ to a code target $\mathcal{M}$$=$$\mathcal{E}_{Q\!R}(M)$ by an aesthetic QR code encoder $\mathcal{E}_{Q\!R}$ \cite{Cox,ARTUP}, aims to reshuffle the module locations to follow the visual features of the content image $I_c$.
$\mathcal{M}$ is an $m\times m$ matrix consisting of 1 or 0, which marks the ideal color of each module (0/1 means black/white). $\mathcal{L}_{code}^{M_k}$ is defined as
\vspace{-0.2cm}
\begin{equation}
\label{equation:codeloss}
\mathcal{L}_{code}^{M_k} \! =  \mathcal{K}_{M_k} \! \cdot \! \| \mathcal{M}_{M_k} - \mathcal{F}_{M_k}\|^2,
\vspace{-0.1cm}
\end{equation}
where $\mathcal{F}$ is the feature map extracted by SS layer\footnote{$\mathcal{M}$ is an $m \! \times \! m$ binary matrix; $\mathcal{F}$ is an $m\!\times \! m$ feature map; Since $\mathcal{M}$$\in$$\{0,1\}$, and $\mathcal{F}$$\in$$[0,1]$, and they can be compared directly.}, $\mathcal{K}$ is an activation map computed by the competition mechanism, and $\mathcal{K}_{M_k}\!\!\!\in$$\mathcal{K}$ is adopted to activate the sub-code-loss $\mathcal{L}_{code}^{M_k}$.

\begin{figure}[t]
\centering
\vspace{-0.4cm}
\includegraphics[width=3.1 in]{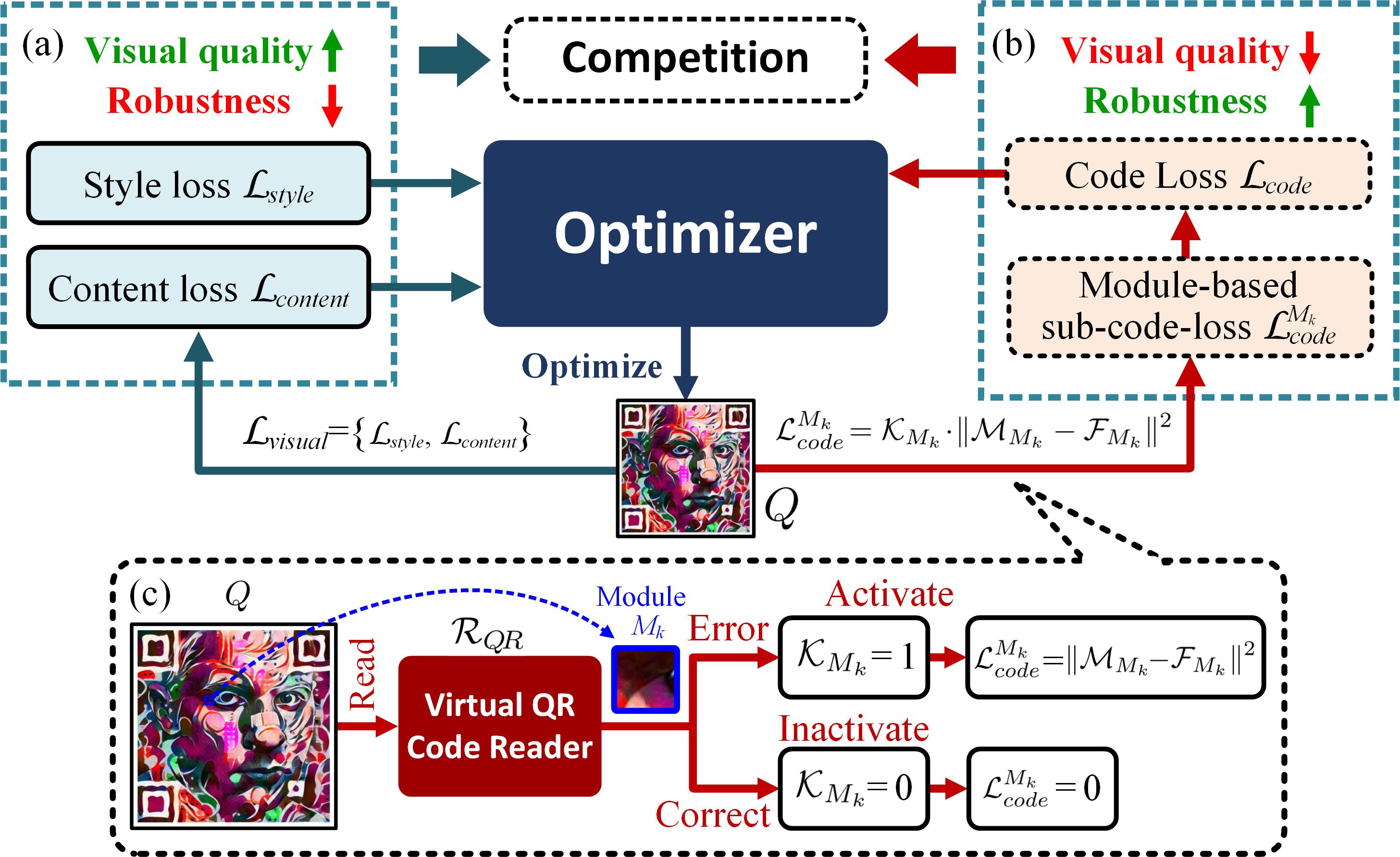}
\vspace{-0.1cm}
\caption{Pipeline of the competition mechanism.~(a)-(b) $\mathcal{L}_{visual}$$=$ $\{$$\mathcal{L}_{style}$,$\mathcal{L}_{content}$$\}$ (or $\mathcal{L}_{code}$) optimizes the visual quality (or robustness) and compromises the robustness (or visual quality), and these losses compete to optimize $Q$. (d) In each iteration, $\mathcal{R}_{Q\!R}$ read all modules, and activate (or inactivate) the sub-code-losses for error (or correct) modules.}
\vspace{-0.4cm}
\label{fig:competition}
\end{figure}

\textbf{Competition mechanism:} As shown in Fig.~\ref{fig:competition}, the main idea of the competition mechanism is that through controlling the activation map $\mathcal{K}$, $\mathcal{L}_{code}$ tries to make each module scanning-robust and compromises the visual quality [Fig.~\ref{fig:competition}(b)], meanwhile, $\mathcal{L}_{style}$ and $\mathcal{L}_{content}$ try to improve the visual quality of $Q$ and compromises the scanning-robustness [Fig.~\ref{fig:competition}(a)].

Specifically, in each iteration, a virtual QR code reader $\mathcal{R}_{Q\!R}$ reads $Q$ to find out all error modules, and then constructs the activation map $\mathcal{K}$ defined as
\vspace{-0.2cm}
\begin{equation}
\label{equation:K}
\mathcal{K}_{M_k}=\left\{
\begin{aligned}
&1 , \ \ \mathrm{if} \ \  \mathcal{R}_{Q\!R}(Q_{M_k}) \oplus  \mathcal{M}_{M_k} = 1 \\
&0 , \ \ \mathrm{if} \ \ \mathcal{R}_{Q\!R}(Q_{M_k}) \oplus  \mathcal{M}_{M_k} = 0
\end{aligned}
\right.,
\vspace{-0.2cm}
\end{equation}
where $\mathcal{M}_{M_k}$ is defined in eq.(\ref{equation:codeloss}), and $\mathcal{R}_{Q\!R}(Q_{M_k})$ is the reading result of the $k$-th module $Q_{M_k}$ of $Q$.

Combining eq.(\ref{equation:totalloss}) and (\ref{equation:codelosstotal})-(\ref{equation:K}), if a module $Q_{M_k}$ is correct (robust), then $\mathcal{K}_{M_k}\!$$\leftarrow$$0$, $\mathcal{L}_{code}^{M_k}$$\leftarrow$$0$, and our model will try the best to optimize $\mathcal{L}_{visual}$$=$$\{\mathcal{L}_{style},\mathcal{L}_{content}\}$, to improve the style and content feature.~Afterwards, if these modifications make $Q_{M_k}$ error, then $\mathcal{K}_{M_k}\!$$\leftarrow$$1$, and $\mathcal{L}_{code}^{M_k}$ will be activated to optimize the robustness of ${Q}_{M_k}$.
As shown in Fig.~\ref{fig:competition}(e), the competition between $\mathcal{L}_{code}$ and $\mathcal{L}_{visual}$ will make the final output $Q$ reach a stable state with preserving both the artistic style and the robustness.

\subsection{Virtual QR Code Reader}
\vspace{-0.1cm}
\label{section:reader}
In this subsection, following the binarization theory of a QR code reader, we design a mechanism to trade-off the visual quality and the scanning-robustness.

\textbf{Binarization of QR codes.}
After a QR code reader scans a QR code $Q$, the sampled colored pixels will be converted to grayscale and binarized to 0 or 1 by a threshold $\mathcal{T}$, defined as
\vspace{-0.1cm}
\begin{equation}
\label{equation:binary}
Q_{M_k}^b= \xi(Q_{M_k},\mathcal{T} \ \! )=\left\{
\begin{aligned}
&0 , \ \ \mathrm{if} \ \  Q_{M_k} <  \mathcal{T} \\
&1 , \ \ \mathrm{if} \ \ Q_{M_k} \geqslant  \mathcal{T}
\end{aligned}
\right.,
\vspace{-0.2cm}
\end{equation}
where $Q_{M_k}^b$ is the binarized result of module $Q_{M_k}$.

According to eq.(\ref{equation:binary}), for black modules, replacing the black color with a lighter color (e.g., dark red) whose gray value below the threshold $\mathcal{T}$, which can still preserve the correct module data (white modules are similar). This manner can make the color of the stylized result more similar to the style target that has a higher visual quality.

\textbf{Virtual QR Code Reader:}
the virtual QR code reader $\mathcal{R}_{Q\!R}$ is designed to discriminate the correctness of each module of $Q$.~Combining eq.(\ref{equation:K}) and (\ref{equation:binary}), the binarization results of $\mathcal{R}_{Q\!R}$ are computed by
\begin{equation}
\label{equation:RQR}
\mathcal{R}_{Q\!R}(Q_{M_k}) =\left\{
\begin{aligned}
&0 , \ \ \  \mathrm{if} \ \ \mathcal{M}_{M_k} \!\! = \! 0  \ \  \mathrm{and} \ \  Q_{M_k} \!\! < \! \mathcal{T}_b \\[-1.5mm]
&1 , \ \ \  \mathrm{if} \ \ \mathcal{M}_{M_k} \!\! = \! 0  \ \  \mathrm{and} \ \    Q_{M_k} \!\! \geqslant \!   \mathcal{T}_b   \\
&0 , \ \ \  \mathrm{if} \ \ \mathcal{M}_{M_k} \!\! = \! 1  \ \  \mathrm{and} \ \    Q_{M_k} \!\! < \!   \mathcal{T}_w   \\[-1.5mm]
&1 , \ \ \  \mathrm{if} \ \ \mathcal{M}_{M_k} \!\! = \! 1  \ \  \mathrm{and} \ \    Q_{M_k} \!\! \geqslant \!   \mathcal{T}_w
\end{aligned}
\right.,
\end{equation}
where $\mathcal{T}_b$ (or $\mathcal{T}_w$) indicates the virtual threshold adopted to binarize black (or white) modules, and $\mathcal{R}_{Q\!R}(Q_{M_k}) \! \oplus \!  \mathcal{M}_{M_k} \!\! =$ 1 (or 0) means module $Q_{M_k}$ is error (or correct).

 According to Sec.~\ref{section:codeloss}, reducing the strictness of error module discrimination, the optimizer will more focus on optimizing the visual quality. Following eq.(\ref{equation:RQR}), the strictness of discrimination is mainly influenced by $\mathcal{T}_b \ \! $ and $ \ \! \mathcal{T}_w$. In grayscale space, we suppose that the distance between the real threshold $\mathcal{T}$ and the virtual threshold $\mathcal{T}_b$/$\mathcal{T}_w$ is proportional to the robustness $\eta $ for black/white modules, and their relationships follow the \emph{Uniform Distribution} as $\eta$$=$$\frac{|\mathcal{T} \! - \! \mathcal{T}_b|}{\mathcal{T}}$$=$$\frac{|\mathcal{T}_w \! - \! \mathcal{T}|}{(255 \! - \! \mathcal{T})}$, where $\mathcal{T}\!$$\in$$[0, 255]$, $\mathcal{T}_b$$\in$$[0, \mathcal{T}]$, $\mathcal{T}_w$$\in$$[\mathcal{T},255]$. Therefore, we can set $\eta$ to control $\mathcal{T}_w$ and $\mathcal{T}_b$, and further trade-off the visual quality and the robustness.

\begin{figure}[t]
\vspace{-0.4cm}
\centering
\includegraphics[width=3.2 in]{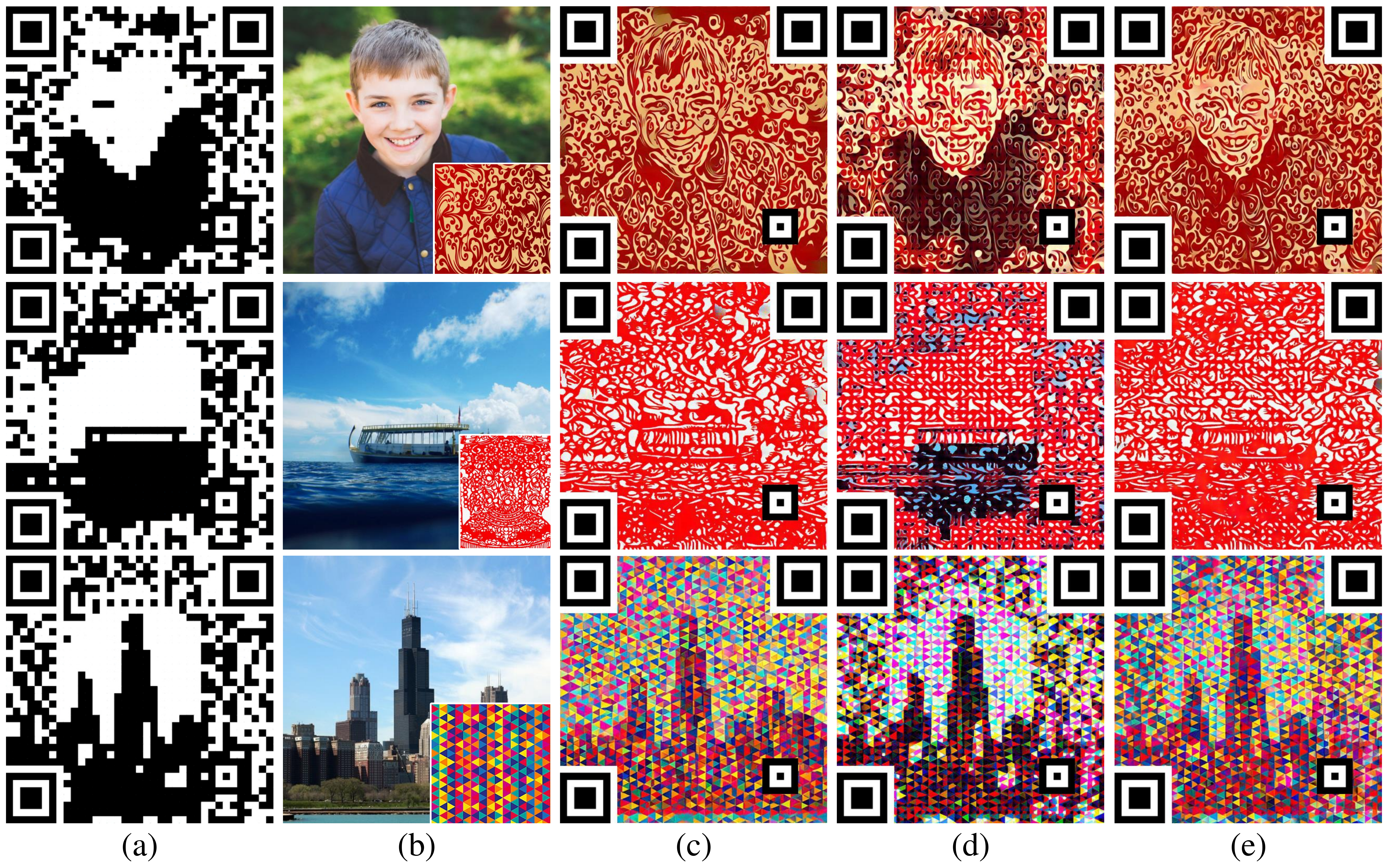}
\vspace{-0.3cm}
\caption{(a) Code target $\mathcal{M}$. (b) Content and style targets. (c) W/O $\mathcal{L}_{code}$. (d) W/O competition mechanism. (e) Ours.  }
\label{fig:ablation}
\vspace{-0.4cm}
\end{figure}

\section{Experiment}
\label{section:experiment}
In the following experiments, we evaluate the performances of our stylized QR codes in two aspects, stylization quality and scanning-robustness.

\subsection{Implementation}
\textbf{Dataset.}~The datasets we used in experiments are divided in two parts, the content image dataset $\mathcal{D}_{c}$, and the style image dataset $\mathcal{D}_{s}$. $\mathcal{D}_{c}$ contains 100 images of size 512$\times$512 with various visual contents (e.g., portrait, cartoon, scenery, animal, logo), and $\mathcal{D}_{s}$ contains 30 images with different artistic styles.

\textbf{Experimental setting.}~We implement our program in PyTorch \cite{pytorch} and all experiments are performed on a computer with a NVIDIA Tesla V100 GPU.~In all experiments to evaluate the scanning-robustness, all QR codes are displayed on a 27-inch, 144Hz, and $3840$$\times$$2160$ IPS-panel monitor screen.~The network adopted to extract features of style and content is VGG-19 \cite{VGG16, Lff} pre-trained on MSCOCO \cite{MScoco}, and the optimizer is Adam.~For all experiments, by default, we set $\lambda_1$$=$$10^{15}\!$, $\lambda_2$$=$$10^7$, $\lambda_3$$=$$10^{20}$ in eq.(\ref{equation:totalloss}), $s$$\in$\{relu1\_2, relu2\_2, relu3\_3, relu4\_3\}, and $c$$\in$\{relu3\_3\} in eq.(\ref{equation:stylecontentloss}), learning rate is $0.001$, robust parameter $\eta$ is $0.6$, and each stylized QR code is output at $10^{4}$ iterations. Moreover, each QR code is generated in version 5 \cite{ISO}, of size $592$$\times$$592$ ($37$$\times$$37$ modules, each module of size $16$$\times$$16$).

\subsection{Stylization Quality}

\begin{figure}[t]
\centering
\vspace{-0.4cm}
\includegraphics[width=3.3 in]{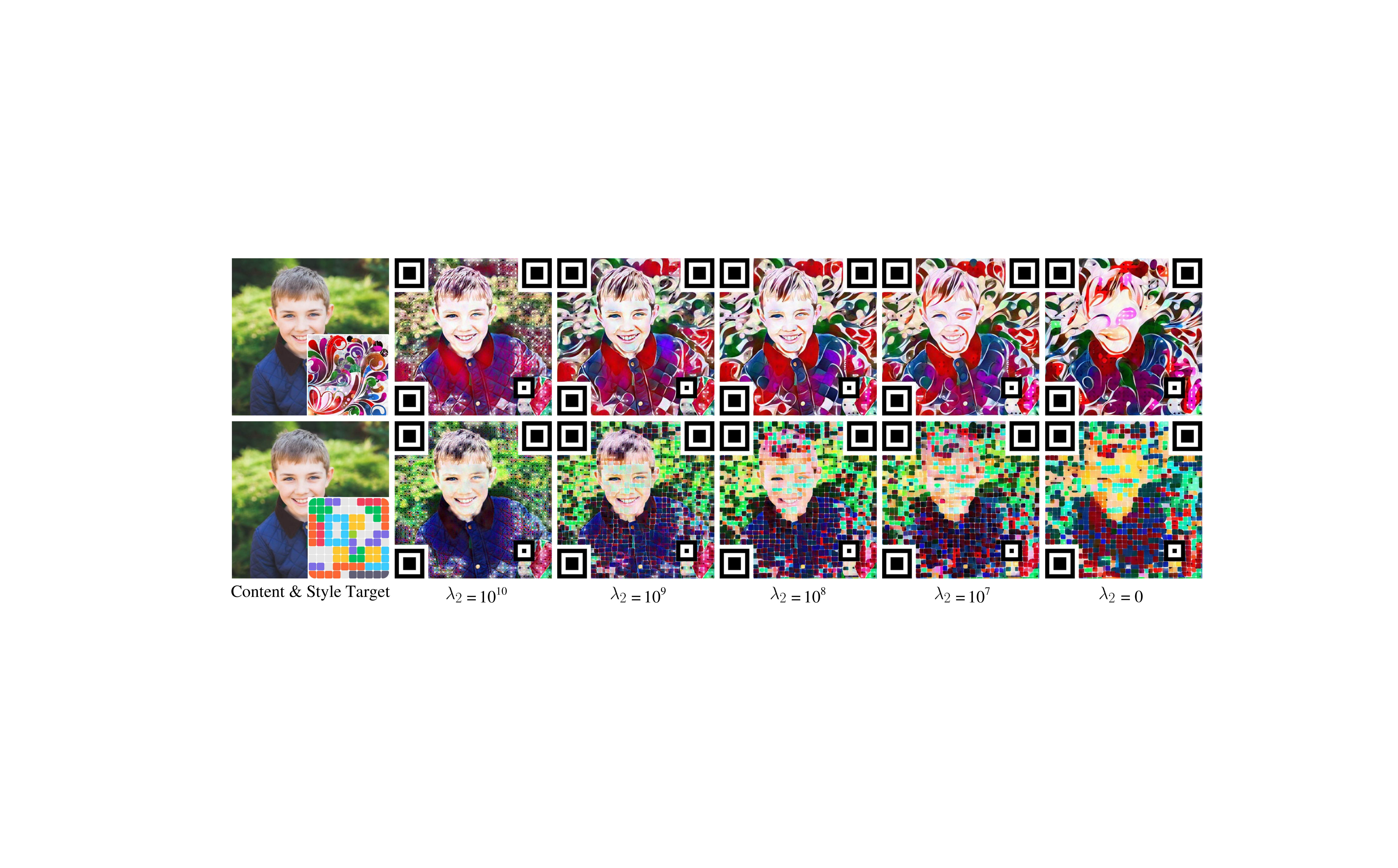}
\vspace{-0.6cm}
\caption{Influences of weights setting.}
\label{fig:contentweightchange}
\end{figure}

\renewcommand\arraystretch{1.4}
\begin{table}[t]\scriptsize
\centering
\caption{Ablation study on our improvements}
\begin{tabular}{m{1.8cm}<{\centering}|m{1.5cm}<{\centering}|m{2cm}<{\centering}|m{1cm}<{\centering}}
\hline
\hline               & W/O $\mathcal{L}_{code}$  & W/O competition mechanism & Ours \\ \hline
\textbf{Visually pleasant}    & \normalsize\checkmark & \scriptsize{\XSolidBrush}                      & \normalsize\checkmark \\
\hline
\textbf{Scanning-robust }     & \scriptsize{\XSolidBrush}  & \normalsize\checkmark                      & \normalsize\checkmark \\
\hline
\hline
\end{tabular}
\label{table:without}
\end{table}
\renewcommand\arraystretch{1}

\textbf{Ablation study.}~Bellow we conduct an ablation study on our improvements.~The produced results under different improvement as shown in Fig.~\ref{fig:ablation}, and the performances are summarized in Tab.~\ref{table:without}. We observe that each of our improvement is essential to produce high-quality results.

The code loss $\mathcal{L}_{code}$ is essential to preserve the scanning-robustness.~As shown in Fig.~\ref{fig:ablation}(c), without $\mathcal{L}_{code}$, the stylized results will lose the functionality of QR codes.~The competition mechanism is essential to balance the scanning-robustness and the visual quality. As shown in Fig.~\ref{fig:ablation}(d), without the mechanism, the modules in stylized results are uncontrolled, and appear as mess black/white patches that are undesirable and visual-unpleasant.

\textbf{Influence of weights setting.}~For the total objective function eq.(\ref{equation:totalloss}), we empirically set the weight of $\mathcal{L}_{code}$ with a larger number to give top priority to keep the scanning-robustness.~With the help of the proposed competition mechanism, a larger $\mathcal{L}_{code}$ does not compromise the visual quality, since the mechanism will inactivate $\mathcal{L}_{code}$ for all robust modules, and $\mathcal{L}_{code}$ will become $0$ when each module is robust.
 In this experiment, we fix $\lambda_1$$=$$10^{15}$, $\lambda_3$$=$$10^{20}$, and only modify the weight $\lambda_2$ of loss $\mathcal{L}_{content}$ to evaluate the visual changing. The comparison results as shown in Fig.~\ref{fig:contentweightchange}, we observe that without compromising the scanning-robustness, fine-tuning weights can effectively trade-off the representation of content and style.

\begin{figure}[t]
\vspace{-0.4cm}
\centering
\includegraphics[width=3.2 in]{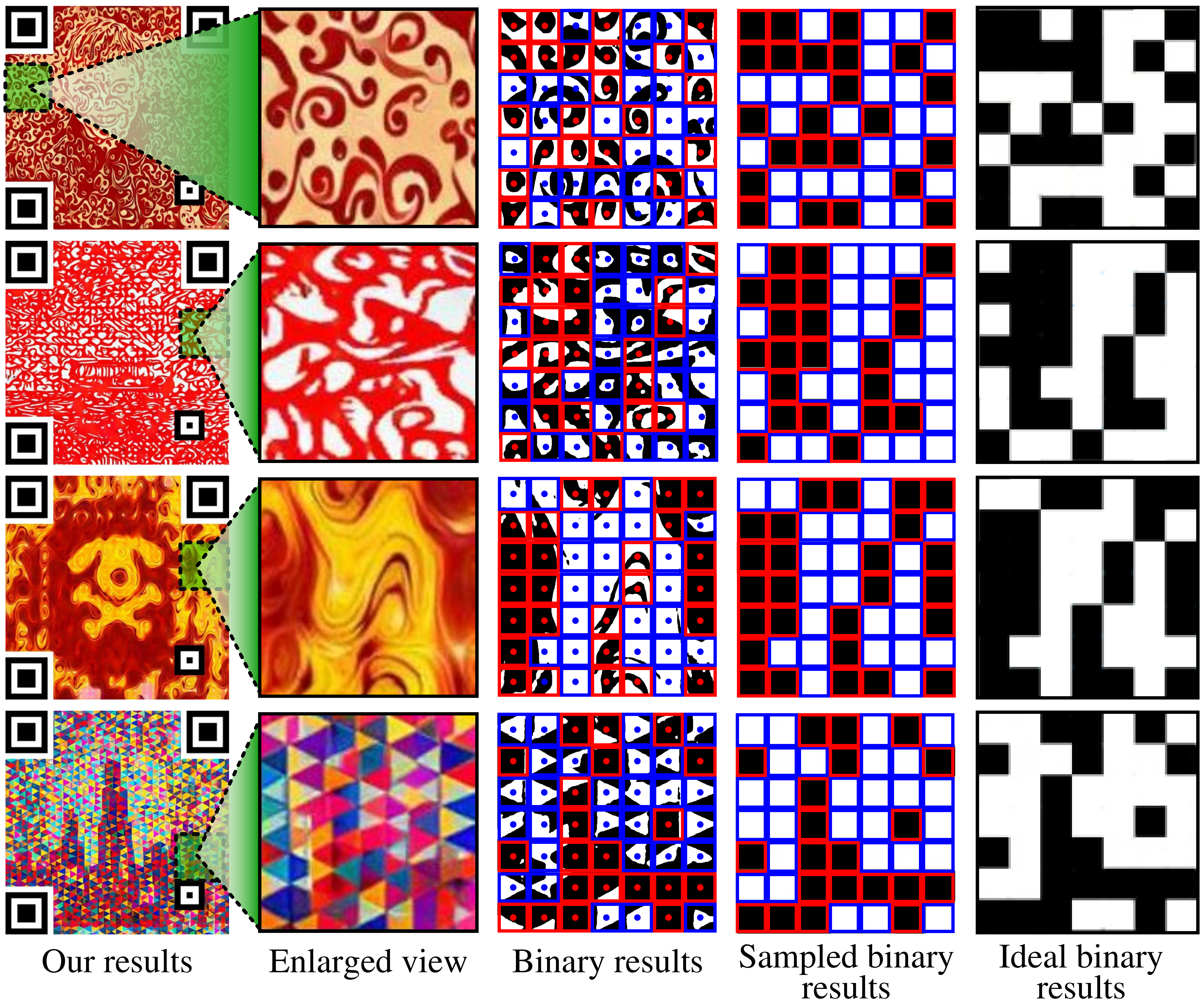}
\vspace{-0.3cm}
\caption{Analysis of preserving scanning-robustness. Red or blue boxes mark the black or white modules, and colored dots mark the pixels sampled by a standard QR code reader.}
\vspace{-0.5cm}
\label{fig:module-visual}
\end{figure}

\begin{figure*}[t]
\vspace{-0.3cm}
\centering
\includegraphics[width=6.3 in]{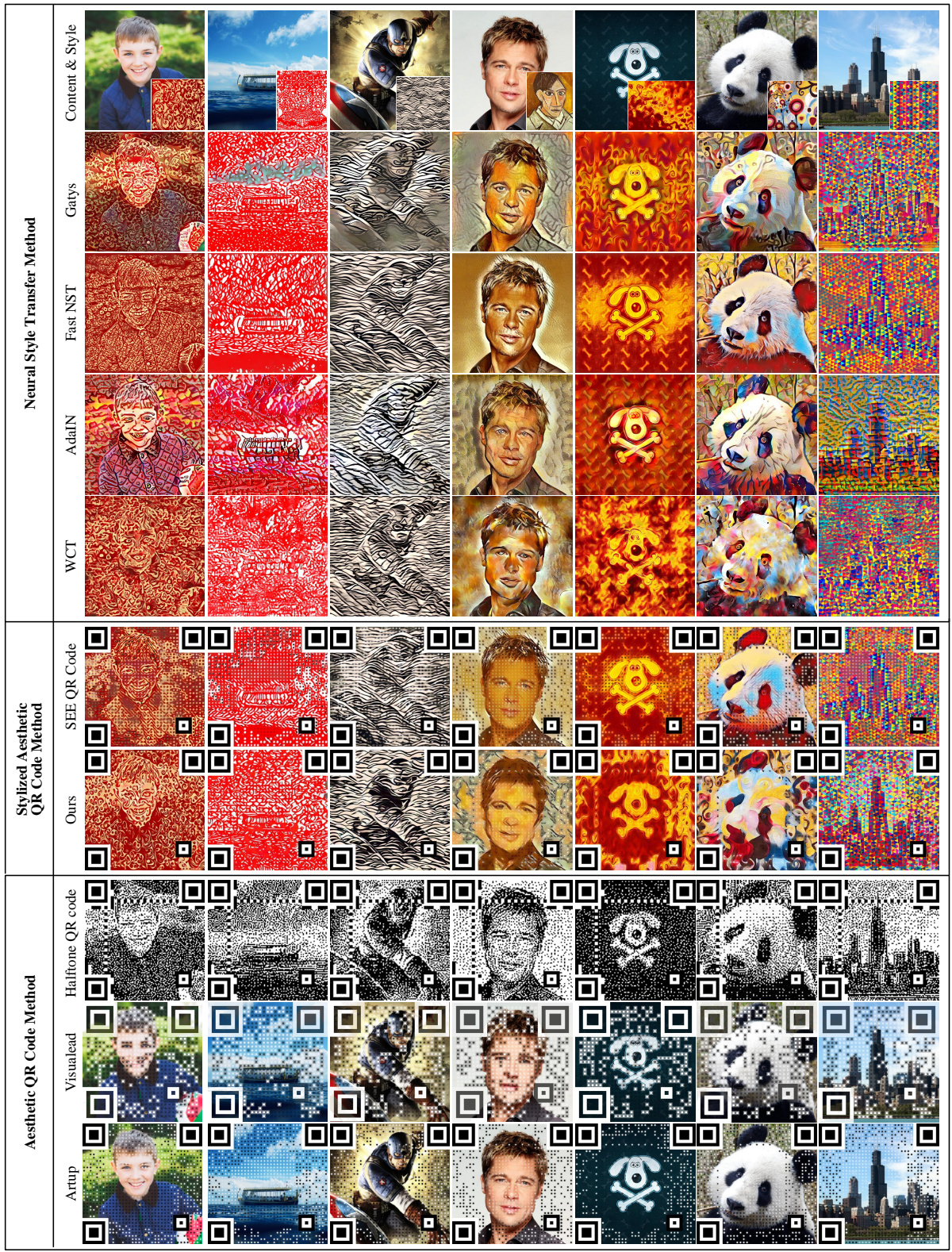}
\vspace{-0.2cm}
\caption{Comparison with previous NST methods and aesthetic QR codes methods. For NST methods, our stylized results achieve a similar stylization quality with them, and outperform them on preserving the functionality of QR codes. For aesthetic QR codes methods, our results are more personalized and diverse and have a higher quality of stylization.}
\label{fig:sota}
\end{figure*}

\textbf{Comparison with other methods.}~We compare our methods with other state-of-the-art NST methods (containing Gatys \emph{et al.} \cite{Gatys}, Fast NST \cite{Lff}, AdaIN \cite{AdaIN}, and WCT \cite{WCT}) and aesthetic QR codes methods (containing SEE QR code \cite{StylizeQR}, Halftone QR code \cite{HF}, Visualead \cite{VS}, and Artup \cite{ARTUP}). The comparison results as shown in Fig.~\ref{fig:sota}, for the compared NST methods, our stylized results achieve a similar stylization quality with them, and outperform them on preserving the functionality of QR codes; for the works of aesthetic QR codes, our methods offer various generating styles that are more personalized, diverse, and artistic. Particularly, for the prior work of stylized QR code \cite{StylizeQR}, we observe that the robustness of their codes relies on the repair by the post-processing algorithm, and the repaired modules appear as some round spots that are visible, undesired, and distracting. Contrarily, all modules in our stylized QR codes are more invisible and well fuse with the entire image.



\subsection{Scanning Robustness}

~Bellow we conduct a series of experiments to evaluate the scanning-robustness of our stylized QR codes in real-world application.

\begin{figure}[t]
\centering
\vspace{-0.4cm}
\includegraphics[width=3.2 in]{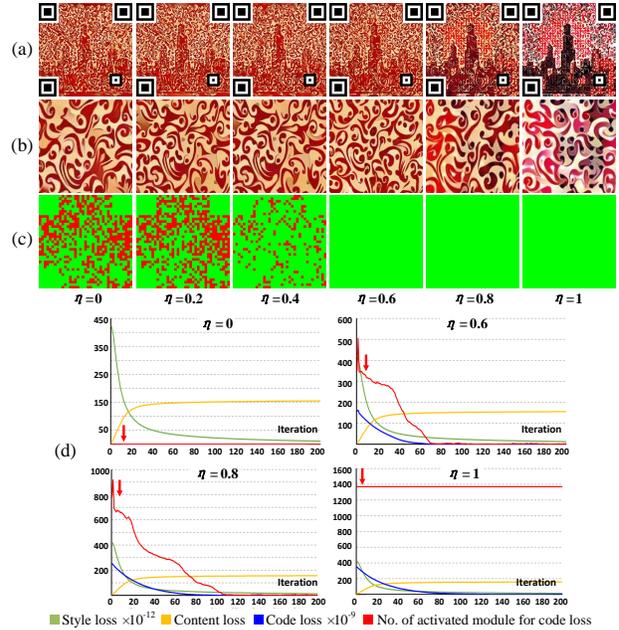}
\vspace{-0.2cm}
\caption{Influences of robustness parameter $\eta$.~(a) Generated results. (b) Enlarged view of (a). (c) Error modules in (a) (marked with red). (d) Influences on losses. $\eta$ is proportional (inversely proportional) to robustness (visual quality).}
\label{fig:robustparam}
\vspace{-0.5cm}
\end{figure}

\textbf{Analysis of preserving scanning-robustness.}~We first analysis and explain why our stylized results can preserve the scanning-robustness. Following the binarization theory of QR codes described in Sec.~\ref{section:reader}, for the sampled pixels of each module, no matter how their color changes, just preserving the same binary results with the ideal colors, can preserve the scanning-robustness.

As shown in Fig.~\ref{fig:module-visual}, for our stylized codes, although the colors and shapes of their modules are invisibly blended with the entire image, and become varied and irregular, the sampled pixels still preserve the same binary results with the ideal QR codes. Therefore, our stylized QR codes can be robustly decoded by a standard QR code reader.

\textbf{Influence of robustness parameter $\bm{\eta}$.}~In Sec.~\ref{section:reader}, the parameter $\eta$ controls the strictness of discriminating error modules (i.e., controls the robustness of modules). Specifically, when we set a higher $\eta$, each module must be blacker/whiter to be classified as a robust module [Fig.~\ref{fig:robustparam}(a)(b)]. For all non-robust modules, their sub-code-losses need to be activated to optimize the robustness.

 In experiment, we randomly select 20 content images and 10 style images from the dataset $\mathcal{D}_{c}$ and $\mathcal{D}_{s}$ respectively, to generate 20 stylized QR codes, and set $\lambda_1$/$\lambda_2$/$\lambda_3$$=$$10^{15}$/$10^{7}$/$10^{20}$ in eq.(\ref{equation:totalloss}).~The experimental results as shown in Fig.~\ref{fig:robustparam}, for visual quality, a higher $\eta$ make all modules' colors become blacker/whiter, which is more visual-unpleasant [Fig.~\ref{fig:robustparam}(a)(b)]; for scanning-robustness, setting a higher $\eta$ can make the network generate more robust stylized codes that have fewer error module [Fig.~\ref{fig:robustparam}(c)]; for losses changes, we obverse that a higher $\eta$ make the model classify more modules as non-robust modules, and further activates these module's sub-code-losses to optimize the robustness [red arrows in Fig.~\ref{fig:robustparam}(d)].
 To sum up, by modifying $\eta$, the proposed method can effectively trade-off the visual quality and robustness, and $\eta$ is proportional to the robustness and inversely proportional to the visual quality.



\textbf{Influences of mobile phone and reader.}~We will evaluate the the influences of different mobile phones and readers as follows. First, we randomly select 10 content images from $\mathcal{D}_{c}$, and 10 style images from $\mathcal{D}_{s}$, to generate a set $\mathcal{D}_{Q}$ of 10 stylized QR codes ($\eta$$=$$0.6$) with resolution 512$\times$512.~Then, each code in $\mathcal{D}_{Q}$ is shown on the screen in three frequently-used sizes, i.e., 3cm$\times$3cm, 5cm$\times$5cm, and 7cm$\times$7cm. At a distance of 20cm, we scan each of these 30 codes using different mobile phones and APPs, and record the average number of successful scanning in 50 scanning-times (a successful scanning is defined as the code can be decoded in 3 seconds).

The experimental results in Tab.~\ref{table:scan-robust} show that the average rates of successful scanning are always greater than $96\%$ (the failure cases can still be decoded, just the decoding time exceeds 3 seconds), which means our stylized codes are robust enough for real-world applications.

\begin{table}[t]\scriptsize
	\centering
	\caption{Average success rates}
\vspace{-0.2cm}
	\begin{tabular}{c|p{1.3cm}<{\centering}|p{1cm}<{\centering}|p{1cm}<{\centering}|p{1cm}<{\centering}}
		\hline
		\hline
		\multicolumn{1}{c|}{\multirow{2}{*}{\begin{minipage}{1.5cm}\vspace{0.4cm}
			{\textbf{Moblie Phone}} \vspace{0.1cm}\end{minipage}}}& \multirow{2}{*}{{\begin{minipage}{0.5cm}\vspace{0.4cm}
			{\textbf{APP}} \vspace{0.1cm}\end{minipage}}}&\multicolumn{3}{p{3.6cm}<{\centering}}{\scriptsize{$\!\!\!$\textbf{Successful scanning / Scanning times}$ \bm{\times 100\%$}}}  \\ \cline{3-5}
		\multicolumn{1}{c|}{}                                &                      &\begin{minipage}{10cm}\vspace{0.1cm}
			{$ \ \ \ \ $(3{cm})$^2$} \vspace{0.1cm}\end{minipage}     &\begin{minipage}{10cm}\vspace{0.1cm}
			{$ \ \ \ \ $(5{cm})$^2$} \vspace{0.1cm}\end{minipage}      & \begin{minipage}{10cm}\vspace{0.1cm}
			{$ \ \ \ \ $(7{cm})$^2$} \vspace{0.1cm}\end{minipage}     \\ \hline \hline
		\multirow{4}{*}{Iphone 11}                            & Twitter               &  $100\%$   &  $100\%$   &  $100\%$  \\ \cline{2-5}
		& Facebook           &  $96\%$   &  $100\%$   &  $100\%$  \\ \cline{2-5}
		& Wechat               &  $100\%$   &  $100\%$   &  $100\%$   \\ \cline{2-5}
		& Alipay      &  $100\%$   &  $100\%$   &  $100\%$   \\ \hline
		\multicolumn{1}{c|}{\multirow{4}{*}{Huawei Mate 20}} & Twitter     &  $100\%$   &  $100\%$   &  $100\%$     \\ \cline{2-5}
		\multicolumn{1}{c|}{}                                & Facebook       &  $100\%$   &  $100\%$   &  $100\%$   \\ \cline{2-5}
		\multicolumn{1}{c|}{}                                & Wechat           &  $96\%$   &  $100\%$   &  $100\%$      \\ \cline{2-5}
		\multicolumn{1}{c|}{}                                & Alipay     &  $100\%$   &  $100\%$   &  $100\%$    \\ \hline
		\multirow{4}{*}{Vivo X30}                       & Twitter             &  $100\%$   &  $100\%$   &  $100\%$     \\ \cline{2-5}
		& Facebook      &  $100\%$   &  $100\%$   &  $100\%$      \\ \cline{2-5}
		& Wechat          &  $98\%$   &  $100\%$   &  $100\%$       \\ \cline{2-5}
		& Alipay   &  $100\%$   &  $100\%$   &  $100\%$     \\ \hline
		\multirow{4}{*}{Xiaomi Note 3}                        & Twitter           &  $100\%$   &  $100\%$   &  $100\%$      \\ \cline{2-5}
		& Facebook      &  $100\%$   &  $100\%$   &  $100\%$      \\ \cline{2-5}
		& Wechat          &  $96\%$   &  $100\%$   &  $100\%$       \\ \cline{2-5}
		& Alipay    &  $98\%$   &  $100\%$   &  $100\%$      \\ \hline
		\hline
	\end{tabular}
	\label{table:scan-robust}
\vspace{-0.2cm}
\end{table}

\textbf{Influences of scanning distances and angles.}~To evaluation the influences of scanning distances and angles, we conduct a comparative experiment on related works of aesthetic QR codes, and our method. First, we generate 5 samples for each related work (including Visualead (VS) \cite{VS}, Halftone QR codes (HF) \cite{HF}, Efficient QR codes (EF) \cite{EF}, Artup (AU) \cite{ARTUP}, SEE QR codes (SE) \cite{StylizeQR}, traditional QR codes, and our results under different $\eta$ setting). Second, each of these codes is fixed to 5cm$\times$5cm and is displayed on a screen, and then we scan each code by a same mobile phone at different distances and angles. The decoding time is recorded as the nearest element of \{0.5s, 1s, 1.5s,...,5s\}, and we rule that the time exceeded 5s means the scanned code is a fail case.

\begin{figure}[t]
\centering
\includegraphics[width=3.2 in]{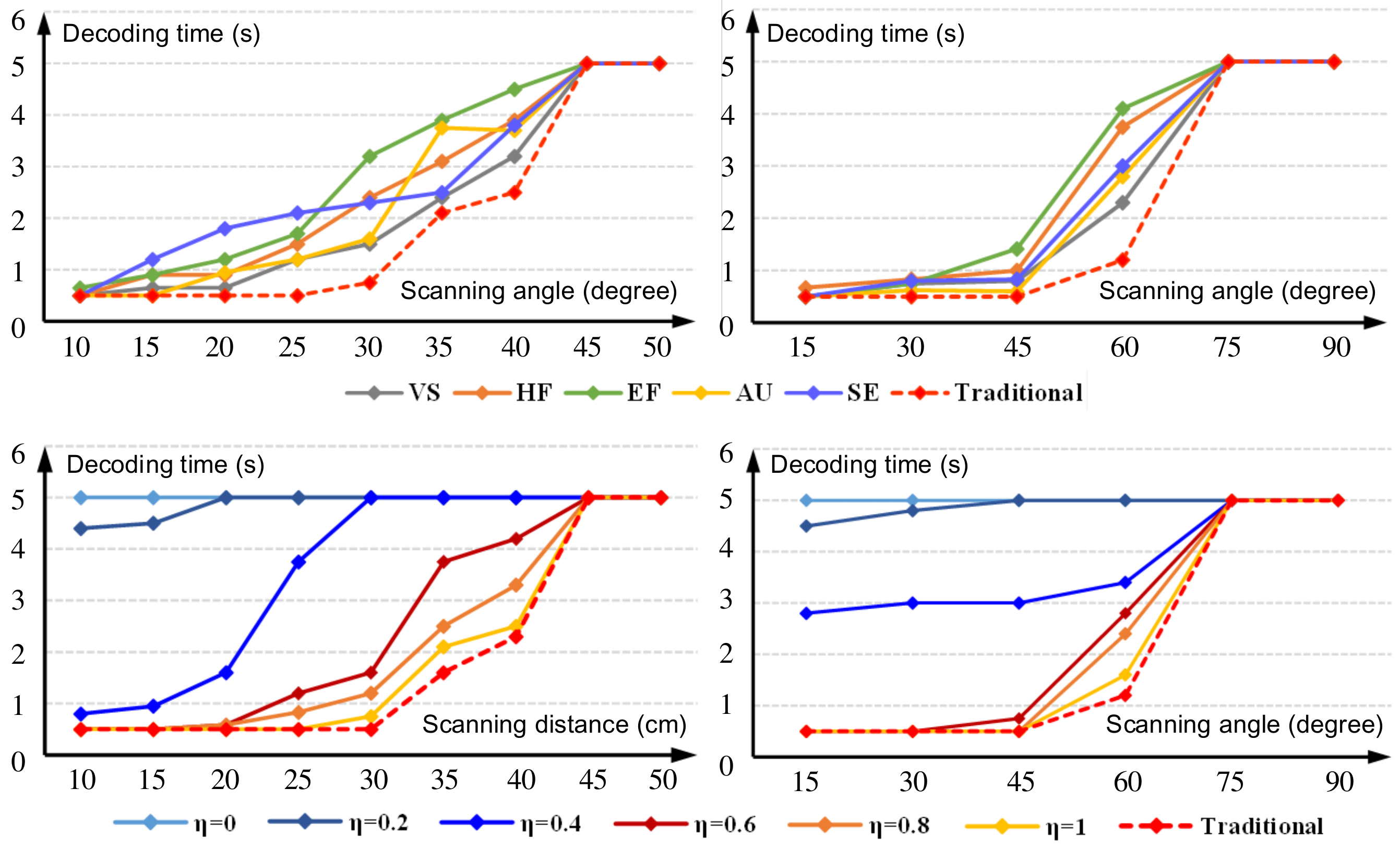}
\vspace{-0.1cm}
\caption{Robustness evaluation on different scanning distances/angles.~\textbf{Upper:} results of related works of aesthetic QR codes. \textbf{Bottom:} our results under different $\eta$ setting. }
\vspace{-0.3cm}
\label{fig:distanceangle}
\end{figure}

The comparison results in Fig.~\ref{fig:distanceangle} show that when $\eta\geqslant 0.6$, the robustness of our stylized codes achieve the similar performance with the related methods.~Moreover, although the robustness of our codes is slightly lower than the traditional QR code, they are robust enough to support the real-world applications.

%
%

\section{Discussion and Conclusion}
In this paper, we propose ArtCode, which can generate the stylized QR codes that are personalized, diverse, and scanning-robust.~To address the challenge that preserving the scanning-robustness after giving such codes style elements, we propose the SS layer, the module-based code loss, and the competition mechanism to improve the performances. Extensive experiments prove that our stylized QR codes have high-quality in both the visual effect and the scanning-robustness, which is able to support the real-world application.

 The proposed ArtCoder is the first end-to-end method that can stylize an image while endowing it with QR code messages. Our current implementation is based on the classical NST method using the iterative optimization (e.g., \cite{Gatys}). ~Although ArtCoder can generate stylized QR codes with good visual quality, compared with the recent fast NST methods (e.g.,\cite{AdaIN, WCT}), ArtCoder still needs improvement in generation speed (it takes an average time of 384.2 seconds to produce a stylized QR code).~How to further increase the generating speed without compromising the scanning-robust and the visual quality will become the vital points to be improved in our future work.

\section{Supplemental Material}
\subsection{Overview}
In this document we provide the following supplementary contents:
\begin{itemize}\setlength{\itemsep}{0pt}
\item an introduction to the fundamental structures of QR codes (Section \ref{sec:QRcode});
\item details about the method to reshuffle the modules of a QR code (Section \ref{sec:GJEP});
\item more generated results of our ArtCoder (Section \ref{sec:samples});
\end{itemize}

\subsection{Basic of QR codes}
\label{sec:QRcode}
\textbf{Fundamental structure of QR Code.} QR codes is a matrix symbology that based on the \emph{Reed-Solomon (RS)} coding rules, whose encoding content is expressed as black or white square \emph{modules} that are divided into the \emph{function patterns} and \emph{encoding region} \cite{ISO}.

\begin{figure}[t]
  \centering

  \includegraphics[width=3.2 in]{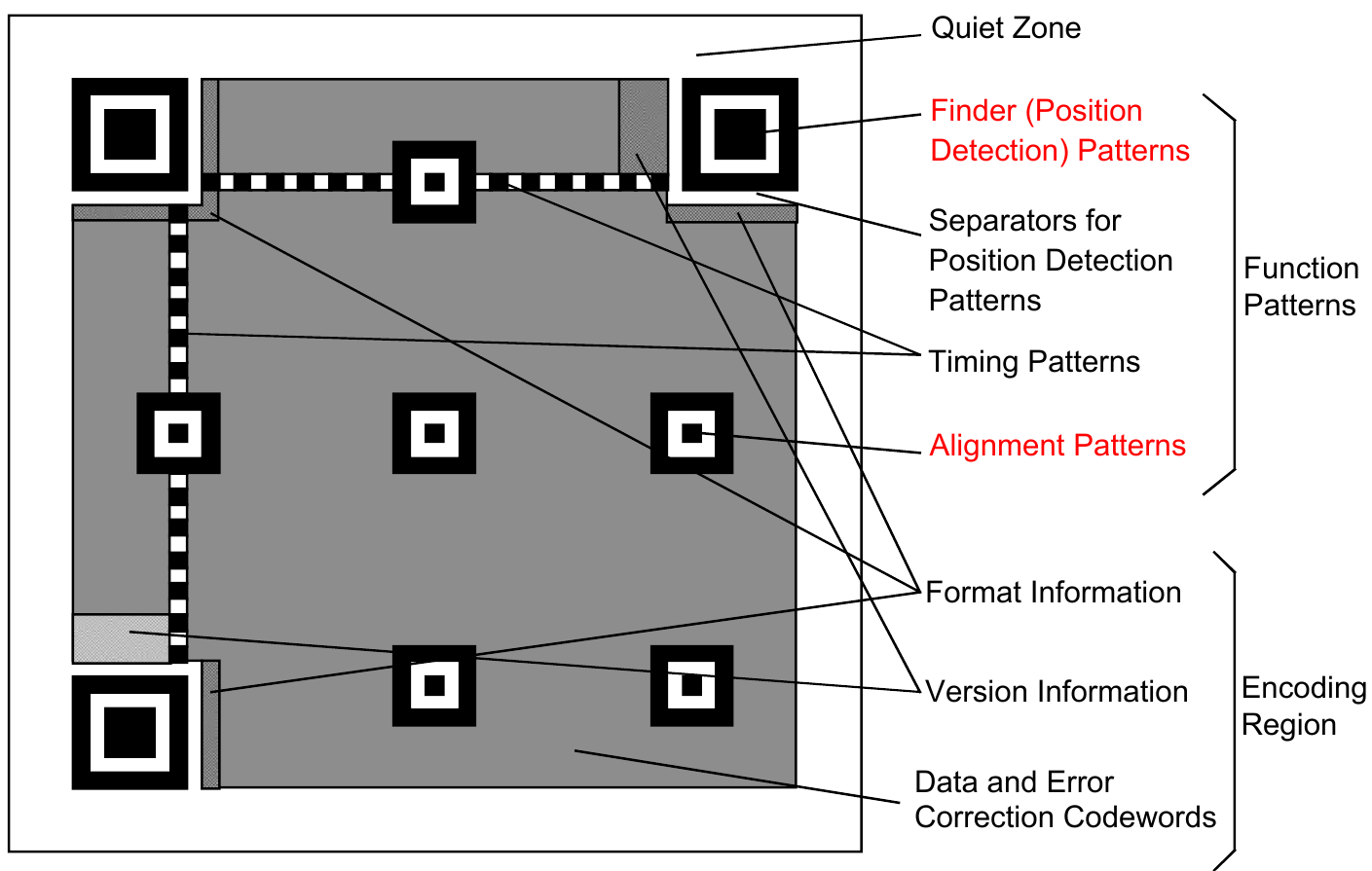}\\
  \caption{Structure of a version 7 QR code.}
  \label{fig:QR}
\end{figure}

\begin{figure}[t]
\centering
\includegraphics[width=3.2 in]{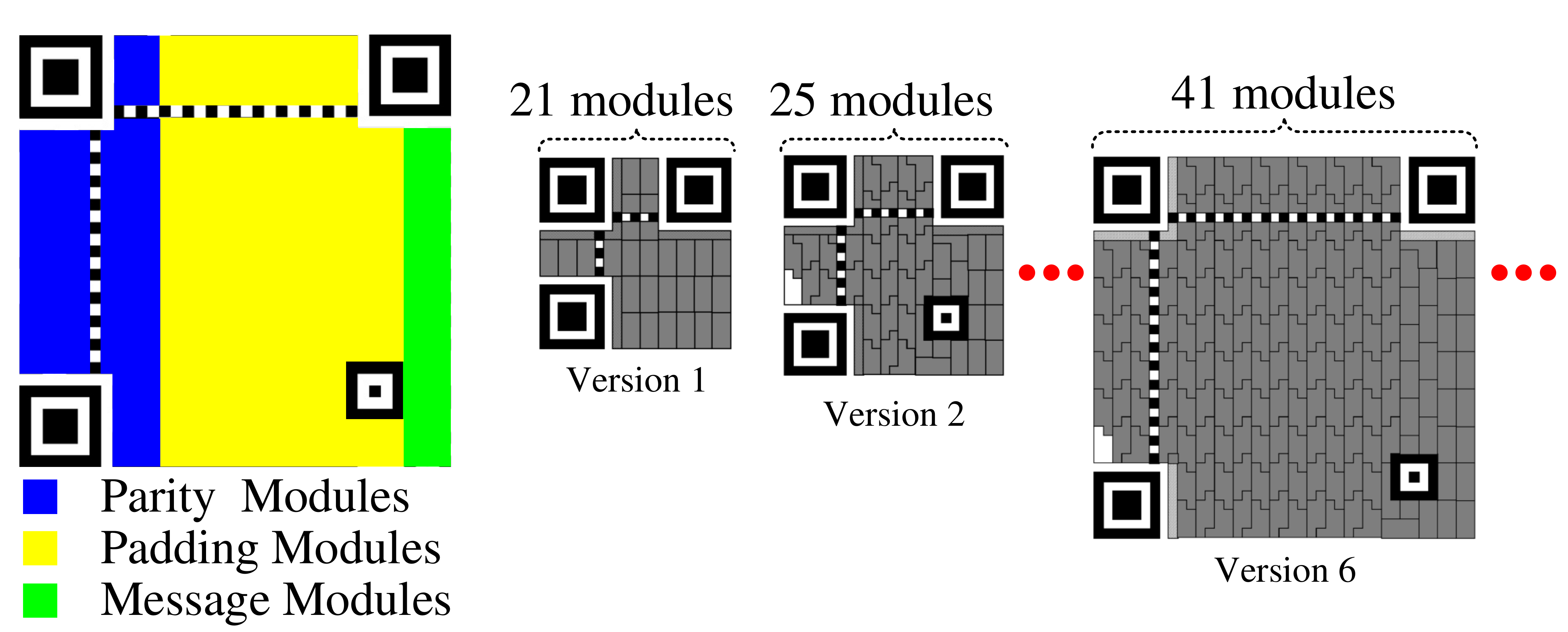}
\caption{\textbf{Left:} QR codes consist of message, padding, and parity modules, a module represents 1-bit data, and a group of modules compose an RS coding block. \textbf{Right:} QR codes of different version number, adhere to the ISO standard \cite{ISO}. For a QR code of version $\mathcal{V}$, each side of it consists of $21+4(\mathcal{V}-1)$ modules. In the main paper, by default, we adopt the QR code of version 5 which is a frequently-used version. }
\label{fig:QRbasic}
\end{figure}

Function patterns include finder, separator, timing patterns, and alignment patterns, which can not be used for the encoding of data. \textbf{\emph{Finder patterns}} and \textbf{\emph{alignment pattern}} are the most important function patterns as shown in Fig.~\ref{fig:QR} (red words), where finder patterns consist of three identical position detection patterns located at the upper-left, upper-right, and lower-left corners of a QR code respectively; an alignment pattern may be viewed as three superimposed concentric squares and is constructed of 5$\times$5 black modules, 3$\times$3 white modules, and a single central black module. In addition, the number of alignment patterns depends on the version of QR code (as shown in the right of Figure \ref{fig:QRbasic}). The encoding region are divided into message/padding/parity modules (as shown in the left of Fig.~\ref{fig:QRbasic}), one module indicates 1-bit data, and a series of modules compose an RS coding block \cite{ISO} which can encode a string of message.

\textbf{Version of QR code .} Following the ISO standard \cite{ISO}, there are 40 sizes of QR Code symbol referred to as Version 1, Version 2, ..., Version 40. Version 1 measures 21$\times$21 modules, Version 2 measures 25$\times$25 modules and so on increasing in steps of 4 modules per side up to Version 40 which measures 177$\times$177 modules (as shown in Fig.~\ref{fig:QRbasic} Right). In the main paper, by default, we adopt the QR code of version 5 which is a frequently-used version.

\begin{figure}[t]
\centering
\includegraphics[width=3.2 in]{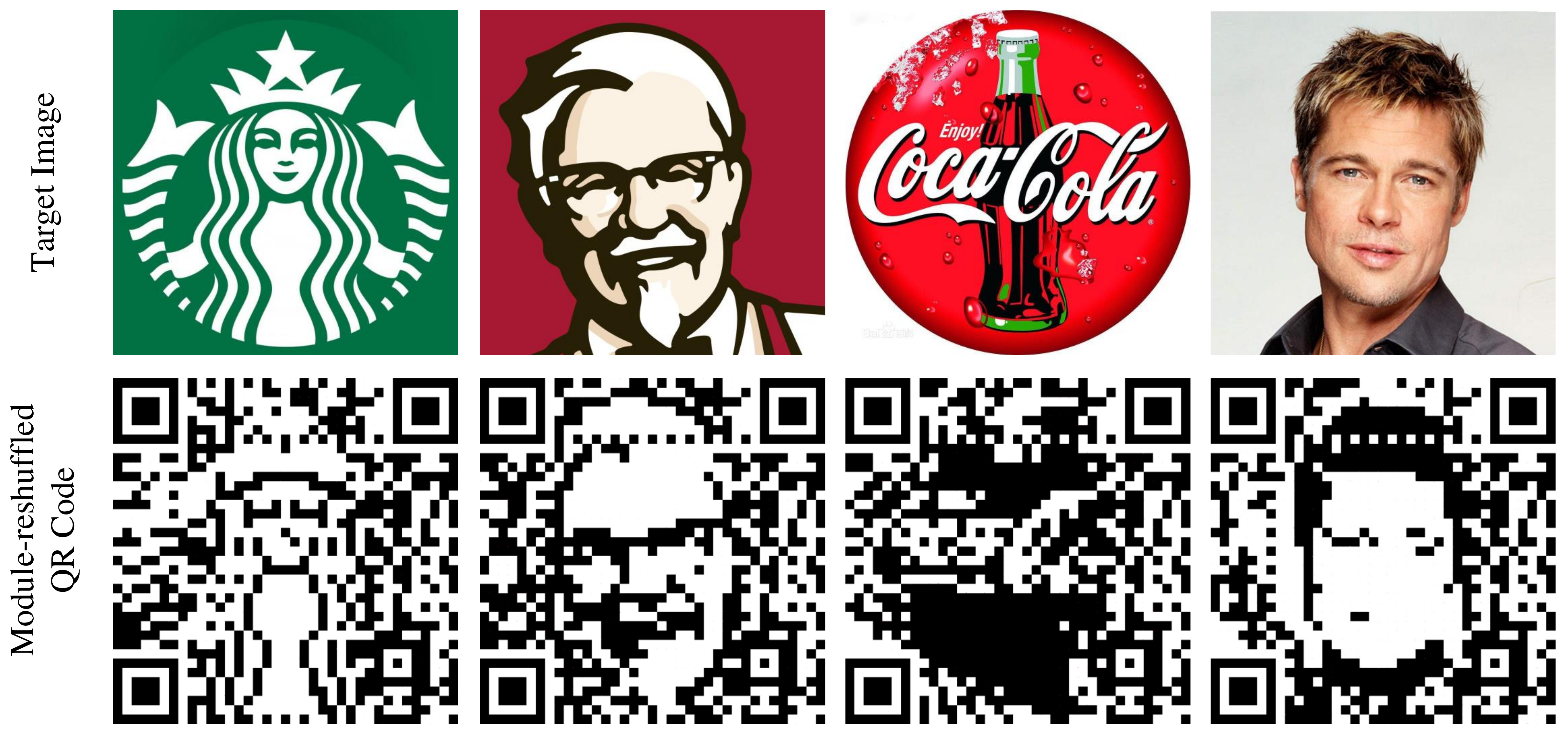}
\caption{Samples of module reshuffle. Gauss-Jordan Elimination Procedure (GJEP) can be employed to reshuffle modules' locations to satisfy the features of target images \cite{Cox,EF,ARTUP}, which gives the QR codes some semantic information. In the main paper, we use the reshuffled QR code as the code target $\mathcal{M}$. Since the module reshuffling is not a key point of our work, we use an off-the-shelf method \cite{ARTUP} to generate the reshuffled QR code.}
\label{fig:GJEP}
\end{figure}

\begin{figure*}[t]
\centering
\includegraphics[width=6.5 in]{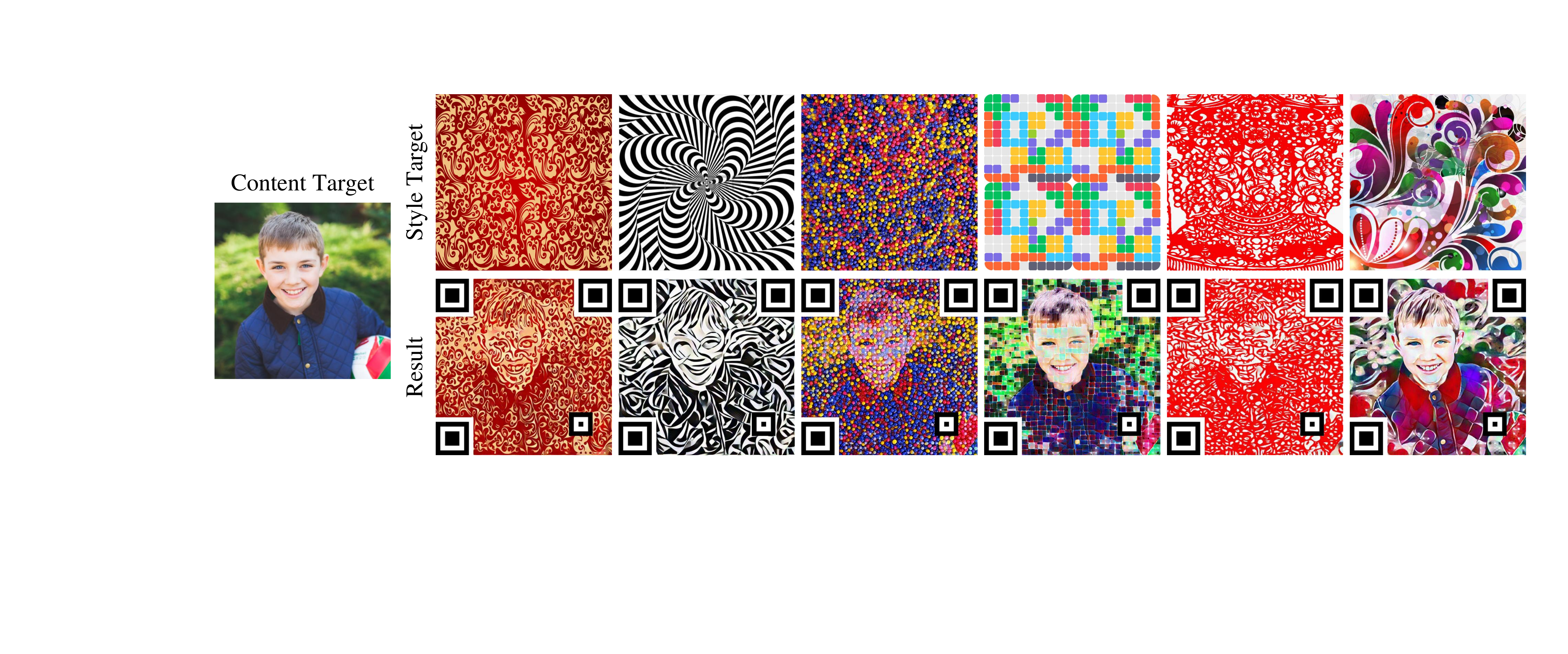}
\caption{Generated samples of stylizing a same content target with different style targets.}
\label{fig:sampel1}
\end{figure*}

\begin{figure*}[t]
\centering
\includegraphics[width=6.5 in]{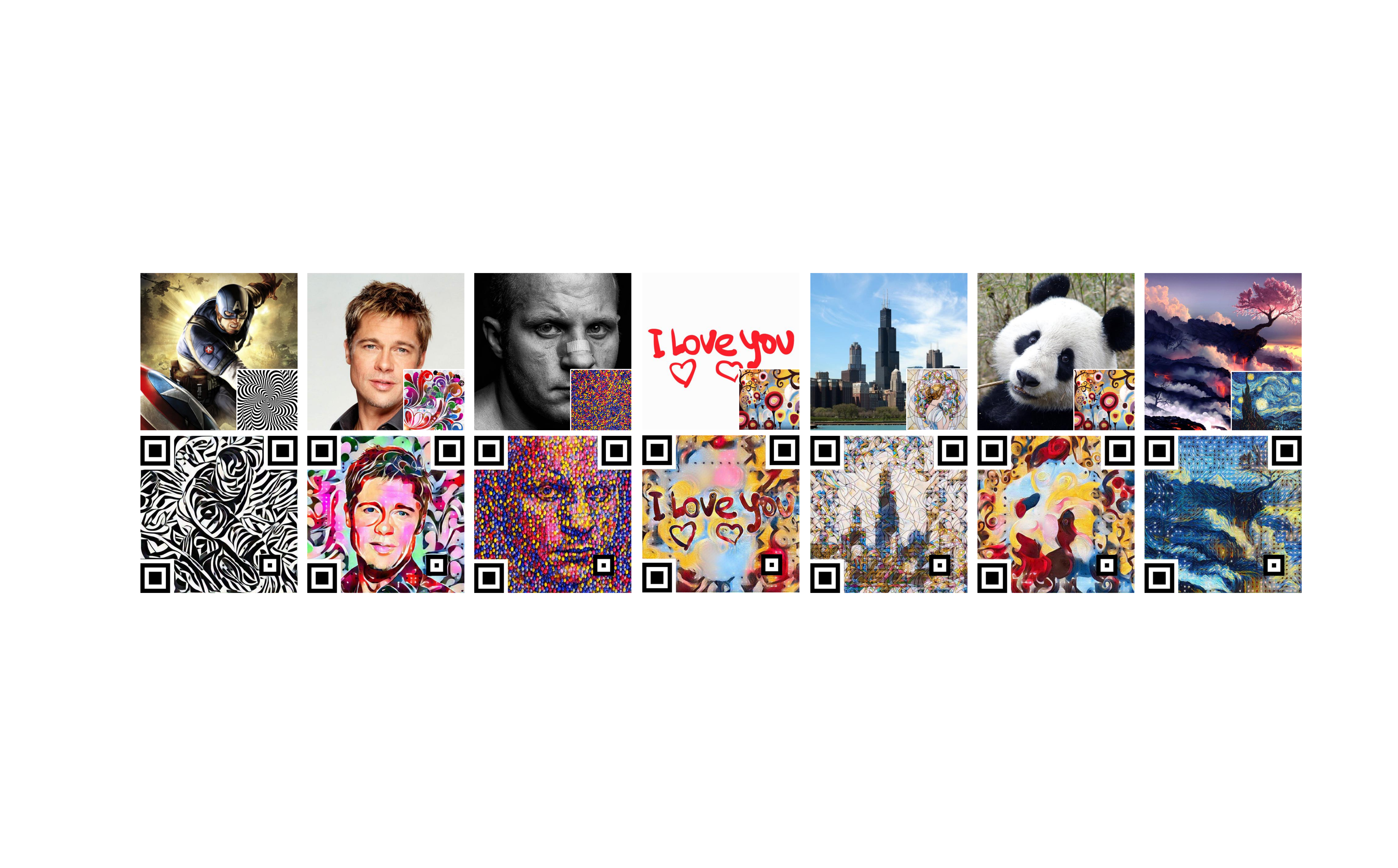}
\caption{Generated samples. The top row shows the content and style pairs and the rest rows present the stylized QR codes.}
\label{fig:sampel2}
\end{figure*}

\subsection{Method of GJEP}
\label{sec:GJEP}
Cox \cite{Cox} have proved that \emph{Gauss-Jordan Elimination Procedure} (GJEP) can be employed to reshuffle QR codes' controllable modules\footnote{The controllable module is defined in \cite{Cox,EF}.} without compromising the scanning-robustness (the reshuffled results as shown in Fig.~\ref{fig:GJEP}). This method uses the vital attribute of RS code that the XOR result of two RS codes is also an RS code \cite{Cox,ARTUP}, so that we can generate a target RS block based on the original RS block and a tailored RS block, for example,
\begin{equation}
\begin{aligned}
&{RS}_{orginal} \!\!\!\! & = [110...000]^m[\underline{1}00101...0]^p[110...01]^t \  \\
&{RS}_{tailored}\!\!\!\! & = [000...000]^m[\underline{1}00000...0]^p[010...10]^t  ,\\
&{RS}_{target}\!\!\!\!   & = [110...000]^m[\underline{0}00101...0]^p[000...11]^t \ \\
\end{aligned}
\label{eq:GJEP}
\end{equation}
where $0/1$ indicates white/black module, the superscript $m / p / t$ indicates the message/padding/parity modules in a RS block. As illustrated in Eq.(\ref{eq:GJEP}), if we intend to change the $k$-th module (underlined) in original RS block, we can create a tailored RS block that $k$-th module is 1, and the rest expect parity modules is 0, the parity modules are computed by the RS coding rules. Then, the XOR result of the original block and the tailored block is the target RS block where the $k$-th module has been changed.

 In the main paper, we use the reshuffled QR code as the code target $\mathcal{M}$. Since the module reshuffling is not a key point of our work, we use an off-the-shelf method \cite{ARTUP} to generate the reshuffled QR code.

\subsection{Generated Sample}
\label{sec:samples}

In this Section, we will show more samples of our generated stylized QR codes with $592\times 592$ resolution. Each generated stylized QR code consists of $37$$\times$$37$ modules, and each module consists of $16$$\times$$16$ pixels.

In Fig.~\ref{fig:sampel1}, we show the generated samples of stylizing a same content target with different style target, and we show some samples generated by different contents and styles in Fig.~\ref{fig:sampel2}.



\begin{thebibliography}{10}
\providecommand{\url}[1]{#1}
\csname url@samestyle\endcsname
\providecommand{\newblock}{\relax}
\providecommand{\bibinfo}[2]{#2}
\providecommand{\BIBentrySTDinterwordspacing}{\spaceskip=0pt\relax}
\providecommand{\BIBentryALTinterwordstretchfactor}{4}
\providecommand{\BIBentryALTinterwordspacing}{\spaceskip=\fontdimen2\font plus
\BIBentryALTinterwordstretchfactor\fontdimen3\font minus
  \fontdimen4\font\relax}
\providecommand{\BIBforeignlanguage}[2]{{%
\expandafter\ifx\csname l@#1\endcsname\relax
\typeout{** WARNING: IEEEtran.bst: No hyphenation pattern has been}%
\typeout{** loaded for the language `#1'. Using the pattern for}%
\typeout{** the default language instead.}%
\else
\language=\csname l@#1\endcsname
\fi
#2}}
\providecommand{\BIBdecl}{\relax}
\BIBdecl

\bibitem{ISO}
ISO, ``Information technology automatic identification and data capture
  techniques code symbology {QR} {Code},'' \emph{Int. Org. Standard.}, Geneva,
  Switzerland, ISO/IEC 18004:~2000.

\bibitem{HF}
H.~K. Chu, C.~S. Chang, R.~R. Lee, and N.~J. Mitra, ``Halftone {QR} codes,''
  \emph{ACM Trans. Graph.}, vol.~32, no.~6, pp. 1--8, 2013.

\bibitem{EF}
S.-S. Lin, M.-C. Hu, C.-H. Lee, and T.-Y. Lee, ``Efficient {QR} code
  beautification with high quality visual content,'' \emph{IEEE Trans.
  Multimedia}, vol.~17, no.~9, pp. 1515--1524, 2015.

\bibitem{ARTUP}
M.~Xu, Q.~Li, J.~Niu, S.~Hao, X.~Liu, W.~Xu, P.~Lv, and B.~Zhou, ``Art-up: A
  novel method for generating scanning-robust aesthetic qr codes,'' \emph{ACM
  Transactions on Multimedia Computing, Communications, and Applications},
  2020.

\bibitem{VS}
\BIBentryALTinterwordspacing
N.~Aliva, U.~Peled, and F.~Itamar. (2012) ``{V}isualead". [Online]. Available:
  \url{http://www.visualead.com/, Accessed on: Mar. 2018.}
\BIBentrySTDinterwordspacing

\bibitem{TS}
Y.~Zhang, S.~Deng, Z.~Liu, and Y.~Wang, ``Aesthetic {QR} codes based on
  two-stage image blending,'' \emph{Springer Int. Publishing}, pp. 183--194,
  2015.

\bibitem{Masic}
Y.-H. Lin, Y.-P. Chang, and J.-L. Wu, ``Appearance-based {QR} code
  beautifier,'' \emph{IEEE Trans. Multimedia}, vol.~15, no.~8, pp. 2198--2207,
  2013.

\bibitem{Gatys}
L.~A. Gatys, A.~S. Ecker, and M.~Bethge, ``Image style transfer using
  convolutional neural networks,'' in \emph{Proc. IEEE Conf. Comput. Vis.
  Pattern Recog.}, 2016, pp. 2414--2423.

\bibitem{Markov}
C.~Li and M.~Wand, ``Combining markov random fields and convolutional neural
  networks for image synthesis,'' in \emph{Proc. IEEE Conf. Comput. Vis.
  Pattern Recog.}, 2016, pp. 2479--2486.

\bibitem{stylebank}
D.~Chen, L.~Yuan, J.~Liao, N.~Yu, and G.~Hua, ``Stylebank: An explicit
  representation for neural image style transfer,'' in \emph{Proc. IEEE Conf.
  Comput. Vis. Pattern Recog.}, 2017.

\bibitem{mangagan}
H.~Su, J.~Niu, X.~Liu, Q.~Li, J.~Cui, and J.~Wan, ``Unpaired photo-to-manga
  translation based on the methodology of manga drawing,'' \emph{arXiv preprint
  arXiv:2004.10634}, 2020.

\bibitem{Lff}
J.~Johnson, A.~Alahi, and F.-F. Li, ``Perceptual losses for real-time style
  transfer and super-resolution,'' in \emph{Proc. Eur. Conf. Comput. Vis.},
  2016, pp. 694--711.

\bibitem{StylizeQR}
M.~Xu, H.~Su, Y.~Li, X.~Li, J.~Liao, J.~Niu, P.~Lv, and B.~Zhou, ``Stylized
  aesthetic qr code,'' \emph{IEEE Transactions on Multimedia}, vol.~21, no.~8,
  pp. 1960--1970, 2019.

\bibitem{reshuff}
S.~Gu, C.~Chen, J.~Liao, and L.~Yuan, ``Arbitrary style transfer with deep
  feature reshuffle,'' in \emph{Proceedings of the IEEE Conference on Computer
  Vision and Pattern Recognition}, 2018, pp. 8222--8231.

\bibitem{SeparatingNST_CVPR2018}
Y.~Zhang, Y.~Zhang, and W.~Cai, ``Separating style and content for generalized
  style transfer,'' in \emph{Proceedings of the IEEE conference on computer
  vision and pattern recognition}, 2018, pp. 8447--8455.

\bibitem{Arbitrary_NST_CVPR2018}
S.~Gu, C.~Chen, J.~Liao, and L.~Yuan, ``Arbitrary style transfer with deep
  feature reshuffle,'' in \emph{Proceedings of the IEEE Conference on Computer
  Vision and Pattern Recognition}, 2018, pp. 8222--8231.

\bibitem{Stroke_NST_ECCV2018}
Y.~Jing, Y.~Liu, Y.~Yang, Z.~Feng, Y.~Yu, D.~Tao, and M.~Song, ``Stroke
  controllable fast style transfer with adaptive receptive fields,'' in
  \emph{Proceedings of the European Conference on Computer Vision (ECCV)},
  2018, pp. 238--254.

\bibitem{text_CVPR2018}
Y.~Men, Z.~Lian, Y.~Tang, and J.~Xiao, ``A common framework for interactive
  texture transfer,'' in \emph{Proceedings of the IEEE Conference on Computer
  Vision and Pattern Recognition}, 2018, pp. 6353--6362.

\bibitem{Arbitraryfast_CVPR2017}
X.~Huang and S.~Belongie, ``Arbitrary style transfer in real-time with adaptive
  instance normalization,'' in \emph{Proceedings of the IEEE International
  Conference on Computer Vision}, 2017, pp. 1501--1510.

\bibitem{NSTCVPR2018}
F.~Shen, S.~Yan, and G.~Zeng, ``Neural style transfer via meta networks,'' in
  \emph{Proceedings of the IEEE Conference on Computer Vision and Pattern
  Recognition}, 2018, pp. 8061--8069.

\bibitem{universalNST_NIPS2017}
Y.~Li, C.~Fang, J.~Yang, Z.~Wang, X.~Lu, and M.-H. Yang, ``Universal style
  transfer via feature transforms,'' in \emph{Advances in neural information
  processing systems}, 2017, pp. 386--396.

\bibitem{Coherent}
D.~Chen, J.~Liao, L.~Yuan, N.~Yu, and G.~Hua, ``Coherent online video style
  transfer,'' in \emph{Proc. Intl. Conf. Computer Vis.}, 2017.

\bibitem{NST_VIDEO_CVPR2017}
H.~Huang, H.~Wang, W.~Luo, L.~Ma, W.~Jiang, X.~Zhu, Z.~Li, and W.~Liu,
  ``Real-time neural style transfer for videos,'' in \emph{Proceedings of the
  IEEE Conference on Computer Vision and Pattern Recognition}, 2017, pp.
  783--791.

\bibitem{stereoscopicNST_CVPR2018}
D.~Chen, L.~Yuan, J.~Liao, N.~Yu, and G.~Hua, ``Stereoscopic neural style
  transfer,'' in \emph{Proceedings of the IEEE Conference on Computer Vision
  and Pattern Recognition}, 2018, pp. 6654--6663.

\bibitem{fastpatch}
T.~Q. Chen and M.~Schmidt, ``Fast patch-based style transfer of arbitrary
  style,'' in \emph{Proc. of NIPS}, 2016.

\bibitem{dia}
J.~Liao, Y.~Yao, L.~Yuan, G.~Hua, and S.~B. Kang., ``Visual attribute transfer
  through deep image analogy,'' \emph{Acm Trans. on Graphics}, vol.~36, no.~4,
  p. 120, 2017.

\bibitem{Cox}
\BIBentryALTinterwordspacing
R.~Cox. (2012) ``{Qartcodes}". [Online]. Available: \url{http://research.
  swtch.com/qart,~Accessed on: Oct. 2017.}
\BIBentrySTDinterwordspacing

\bibitem{VGG16}
K.~Simonyan and A.~Zisserman, ``Very deep convolutional networks for
  large-scale image recognition,'' \emph{Comput. Sci.}, 2014.

\bibitem{ZXing}
\BIBentryALTinterwordspacing
O.~S. (2013) ``{ZXing}". [Online]. Available: \url{https://github.com/zxing/
  zxing, Accessed on: Mar. 2018.}
\BIBentrySTDinterwordspacing

\bibitem{pytorch}
A.~Paszke, S.~Gross, S.~Chintala, G.~Chanan, E.~Yang, Z.~DeVito, Z.~Lin,
  A.~Desmaison, L.~Antiga, and A.~Lerer, ``Automatic differentiation in
  pytorch,'' 2017.

\bibitem{MScoco}
T.~Y. Lin, M.~Maire, S.~Belongie, J.~Hays, P.~Perona, D.~Ramanan, P.~Dollár,
  and C.~L. Zitnick, ``Microsoft {COCO}: common objects in context,'' in
  \emph{Proc. Eur. Conf. Comput. Vis.}, vol. 8693, 2014, pp. 740--755.

\bibitem{AdaIN}
X.~Huang and S.~Belongie, ``Arbitrary style transfer in real-time with adaptive
  instance normalization,'' in \emph{Proceedings of the IEEE International
  Conference on Computer Vision}, 2017, pp. 1501--1510.

\bibitem{WCT}
Y.~Li, C.~Fang, J.~Yang, Z.~Wang, X.~Lu, and M.-H. Yang, ``Universal style
  transfer via feature transforms,'' in \emph{Advances in neural information
  processing systems}, 2017, pp. 386--396.

\end{thebibliography}
\end{document}